%% file: main.tex
\documentclass[acmsmall]{acmart}

\makeatletter
\def\@ACM@checkaffil{
    \if@ACM@instpresent\else
    \ClassWarningNoLine{\@classname}{No institution present for an affiliation}%
    \fi
    \if@ACM@citypresent\else
    \ClassWarningNoLine{\@classname}{No city present for an affiliation}%
    \fi
    \if@ACM@countrypresent\else
        \ClassWarningNoLine{\@classname}{No country present for an affiliation}%
    \fi
}
\makeatother

\usepackage{multirow}
\usepackage{diagbox}
\usepackage{graphicx}
\usepackage{subfigure}
\usepackage{enumitem}
\acmJournal{TIST}
\usepackage{xcolor}
\usepackage{makecell}   

\AtBeginDocument{%
  \providecommand\BibTeX{{%
    \normalfont B\kern-0.5em{\scshape i\kern-0.25em b}\kern-0.8em\TeX}}}

\AtBeginDocument{%
  \providecommand\BibTeX{{%
    \normalfont B\kern-0.5em{\scshape i\kern-0.25em b}\kern-0.8em\TeX}}}

\renewcommand\footnotetextcopyrightpermission[1]{} 
\providecommand{\rev}[1]{#1}
\providecommand{\revmajor}[1]{#1}

\newcommand{\para}[1]{{\vspace{3pt} \bf \noindent #1 \hspace{0pt}}}

\begin{document}

\title{Dynamic Population Distribution Aware Human Trajectory Generation with Diffusion Model}

\begin{abstract}

Human trajectory data is crucial in urban planning, traffic engineering, and public health. However, directly using real-world trajectory data often faces challenges such as privacy concerns, data acquisition costs, and data quality. A practical solution to these challenges is trajectory generation, a method developed to simulate human mobility behaviors. Existing trajectory generation methods mainly focus on capturing individual movement patterns but often overlook the influence of population distribution on trajectory generation. In reality, dynamic population distribution reflects changes in population density across different regions, significantly impacting individual mobility behavior. Thus, we propose a novel trajectory generation framework based on a diffusion model, which integrates the dynamic population distribution constraints to guide high-fidelity generation outcomes. Specifically, we construct a spatial graph to enhance the spatial correlation of trajectories. Then, we design a dynamic population distribution aware denoising network to capture the spatiotemporal dependencies of human mobility behavior as well as the impact of population distribution in the denoising
process. Extensive experiments show that the trajectories generated by our model can resemble real-world trajectories in terms of some critical statistical metrics, outperforming state-of-the-art algorithms by over 54\%. 
\end{abstract}


\author{Qingyue Long}
\affiliation{%
\institution{Department of Electronic Engineering, Beijing National Research Center for Information Science and
Technology (BNRist), Tsinghua University}
\country{China}}
\email{longqy21@mails.tsinghua.edu.cn}

\author{Can Rong}
\affiliation{%
\institution{Department of Electronic Engineering, Beijing National Research Center for Information Science and
Technology (BNRist), Tsinghua University}
\country{China}}
\email{rc20@mails.tsinghua.edu.cn}

\author{Tong Li}
\affiliation{%
\institution{Department of Electronic Engineering, Beijing National Research Center for Information Science and
Technology (BNRist), Tsinghua University}
\country{China}}
\email{tongli@mail.tsinghua.edu.cn}

\author{Yong Li}
\affiliation{%
\institution{Department of Electronic Engineering, Beijing National Research Center for Information Science and
Technology (BNRist), Tsinghua University}
\country{China}}
\email{liyong07@tsinghua.edu.cn}

\ccsdesc[500]{Information systems~Spatial-temporal systems}
\ccsdesc[500]{Computing methodologies~Modeling methodologies}
\ccsdesc[500]{Computing methodologies~Neural networks}

\keywords{Trajectory generation, diffusion models, population distribution}

\maketitle

\input{introduction}

\input{preliminaries}

\input{methods}

\input{experiments}

\input{relatedwork}

\section{CONCLUSION}
In this paper, we propose a novel trajectory generation method that can satisfy the dynamic population distribution. To effectively capture the spatial dependencies of trajectories, we construct a spatial graph and incorporate the geographical proximity into location representations through graph embedding. To ensure the generated trajectories match actual population distributions, we use population distribution as a critical condition in the denoising process and build it into the loss function as a constraint. Extensive experiments show the superiority of our model design and the utility of generated trajectory data.
In the future, we aim to extend our framework by incorporating semantic information, such as the functionality of locations visited by users.



\section*{Acknowledgment}
This work is supported in part by the National Natural Science Foundation of China under 62471277. This work is also supported in part by the National Key Research and Development Program of China under 2023YFB2904804.

\newpage


\clearpage
\bibliographystyle{ACM-Reference-Format}
\bibliography{sample-base.bib}

\end{document}

%% file: introduction.tex
\section{Introduction}
Due to the unique application value and significance, human mobility data is widely used in various fields, including but not limited to urban planning, traffic engineering, and public health~\cite{lu2020inferring,feng2021context,silva2018discovering}. However, despite the irreplaceable value of real trajectory data, it is often not directly usable in practical applications. 
Firstly, using real trajectory data directly can lead to serious privacy breaches~\cite{xu2017trajectory,vlachos2015data}. Even after de-identification or anonymization, there is still a possibility of re-identifying personal information through data fusion and inference~\cite{feng2020pmf,benarous2022synthesis}. Secondly, acquiring real trajectory data is often associated with high costs, such as the need for large-scale deployment of sensor devices or requiring active participation and cooperation from users. Additionally, real data usually contains noise, leading to lower accuracy in subsequent applications. For these reasons, simulating human mobility behavior to generate realistic and high-quality trajectory data has become crucial in downstream applications, attracting widespread attention from academia and industry.

Early trajectory generation methods use traditional machine learning models or statistical methods to generate human trajectories~\cite{jin2022trajectory,han2023st,xu2015map}. For example, methods based on Gaussian Mixture Models (GMM)~\cite{reynolds2009gaussian}, Markov Chains~\cite{ching2006markov}, and Conditional Random Fields (CRF)~\cite{sutton2012introduction} are employed. These methods generated new trajectory samples by modeling the statistical characteristics and patterns of trajectories. However, human movement demonstrates intricate sequential transitions across locations, which can be high-order and time-dependent.
Recent advancements in deep generative AI models offer a promising approach to simulating complex human mobility behavior. For example, Variational Autoencoders (VAE)~\cite{doersch2016tutorial} and Generative Adversarial Networks (GANs)~\cite{creswell2018generative} can learn high-dimensional representations and latent spaces of trajectory data, enabling them to generate new trajectory samples. These deep generative models demonstrate significantly higher accuracy compared to statistical models. Although deep generative models for trajectory generation perform well in simulating individual behaviors, they primarily focus on capturing complex individual movement patterns, often overlooking the impact of population distribution on individual mobility. \rev{In fact, dynamic population distributions impose significant constraints on individual behavior. Some trajectory prediction studies have attempted to incorporate collective information to enhance individual prediction~\cite{bontorin2025mixing}. Prediction refers to estimating future phenomena or behaviors based on historical information. In contrast, our generation task aims to produce datasets of individual mobility behaviors consistent with population distribution dynamics. In other words, we sample and generate plausible individual trajectories under given or possible population distribution variations, rather than forecasting an expected future. However, there is still a problem in effectively describing the complex role of population distribution on individual trajectories in trajectory generation tasks.}

\begin{figure*}[t]
\centering
\subfigure[The impact of dynamic population distribution on individual trajectories.]{\includegraphics[width=.47\textwidth,height=5cm]{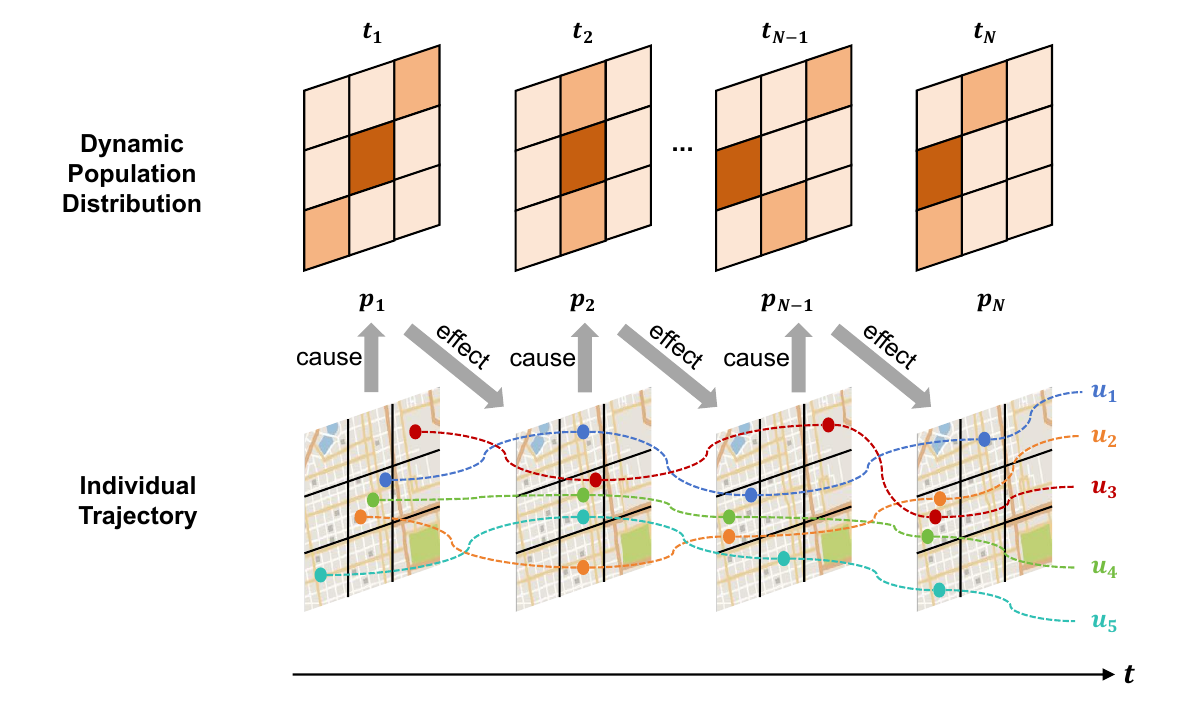}}
\subfigure[The normalized dynamic population and trajectory count for a residential community in Nanchang.]{\includegraphics[width=.47\textwidth,height=5cm]{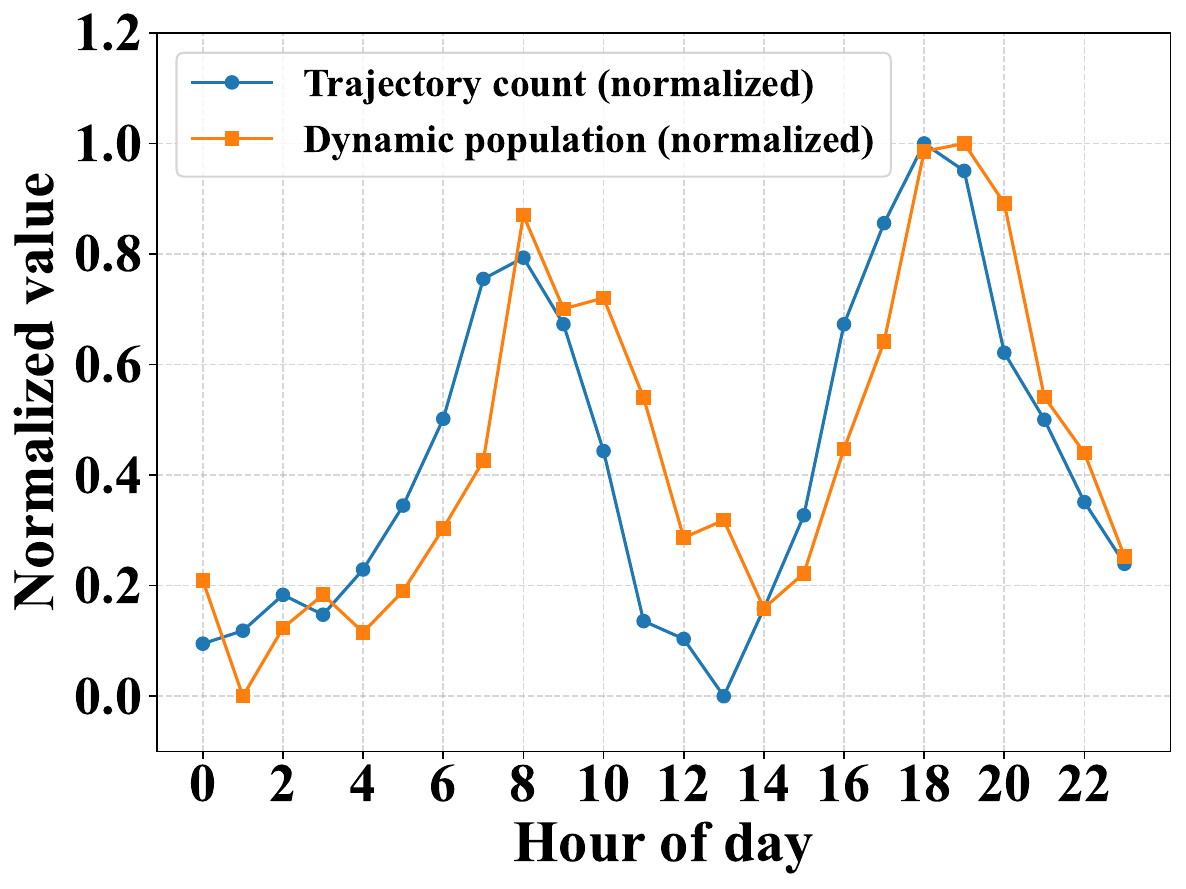}}
\caption{Relationship between dynamic population distribution and individual trajectories.}
\label{fig:trajectory}
\end{figure*}

To address this issue, we propose a novel dynamic population distribution aware trajectory generation method that generates high-quality trajectories while successfully capturing the intricate behaviors of real-world activities.
For the following reasons, we have developed trajectory generation using a diffusion model:
1) The diffusion model performs exceptionally well, yielding impressive outcomes in diverse sequential generation tasks~\cite{zhu2023diffusion, yuan2023spatio,li2023diffurec};
2) Trajectory data typically exhibits high continuity and temporal dependency. Diffusion models produce smooth and coherent trajectories that capture subtle changes by gradually reconstructing the data from random noise. In contrast, GANs and VAEs might face limitations in generating such continuous and smooth trajectories, especially with long sequences or complex dynamic systems~\cite{yang2023diffusion};
3) Trajectory generation often requires the generation of specific trajectories based on certain conditions. Diffusion models can effectively integrate various conditional information to provide fine-grained control for trajectory generation. Although GAN and VAE can also be trained conditionally, the diffusion model is more accurate and flexible in responding to conditions while maintaining the quality of generation~\cite{von2023fabric}.
However, training a population distribution aware trajectory generation model is difficult due to the following challenges:
\begin{itemize}[leftmargin=*]
\item \textbf{Complex spatial dependencies of human mobility behavior.} Individuals' movements within geographical spaces are influenced by the surrounding environmental factors. Urban planning often adopts the "15-minute city" concept, aiming to enable residents to reach various facilities required for daily life within 15 minutes. Under this planning concept, human mobility is significantly influenced by the geographical proximity between locations, with people more likely to frequently visit locations that are geographically close, such as supermarkets near their homes. However, most of the existing studies focus on the sequential relationships of trajectories, ignoring important geospatial information~\cite{feng2018deepmove,qin2018spatio,gao2022trajectory}.
\item \textbf{There are scale differences between macro population distribution and micro individual mobility.} Dynamic population distribution reflects the macroscopic changes in group dynamics, while individual trajectory data reflects the microscopic patterns of individual movements. \rev{Figure~\ref{fig:trajectory} (a) shows a feedback loop: aggregated trajectories shape the population distribution, which in turn guides subsequent movements. Figure~\ref{fig:trajectory} (b) confirms a strong temporal correlation between trajectory count and population in a Nanchang community. Thus, dynamic population distribution is an informative prior for individual mobility generation.} However, the current methods used to generate individual trajectories often fail to capture the impact of population distribution adequately~\cite{feng2020learning, feng2020pmf, choi2021trajgail}.
\end{itemize}

We propose a dynamic population distribution aware trajectory generation method based on the diffusion model to address the challenges mentioned above. This method includes two components: 1) the spatial enhancement module and 2) the diffusion and denoising module based on dynamic population distribution perception.
To solve the first challenge, we construct a spatial graph based on geographic proximity between locations in the spatial information enhancement module. We utilize specific graph embedding techniques to learn embeddings of locations, enabling accurate capture of the interrelationships and geographical proximity between them.
To address the second challenge, we leverage the refined modeling capabilities of the diffusion model, incorporating population distribution conditions throughout the denoising process. Subsequently, we design a population distribution aware loss function, ensuring that the probability of individuals choosing locations aligns with the population distribution.

Overall, the contributions of our paper can be summarized as follows:
\begin{itemize}[leftmargin=*]
    \item We propose a dynamic population distribution aware trajectory generation method, which can generate high-quality trajectories consistent with the population distribution.
    \item We employ the diffusion model to generate trajectories based on population distribution. On the one hand, we design a spatial enhancement module that effectively captures the geographical proximity between locations. On the other hand, we integrate dynamic population distribution into the denoising process and use it as a constraint in the loss function, enabling that the generated trajectories conform to the population distribution.
    \item Our model has been extensively evaluated on two real mobility datasets. The results show that our model has superior performance on many metrics (especially relevant metrics regarding population distribution) with an average improvement of more than 50\%. Further studies also validate the usefulness of our model.
\end{itemize}

%% file: preliminaries.tex
\section{Preliminaries}
\subsection{Problem Definition}
\para{Definition 1 (Mobility Trajectory).} A series of spatial-temporal points $R=\{r_1,...,r_N\}$ make up the mobility trajectory of user $u$. Each spatial-temporal point $r_i$ may be defined as $(l_i,t_i)$, $t_i$ represents the timestamp of the $i$-th visit, and $l_i$ indicates the location information, which is often given as a region ID.

\para{Definition 2 (Population Distribution).} $P=\{p_1,...,p_N\}$ represents the population distribution in both temporal and spatial dimensions, where $N$ is the temporal dimension (e.g., the number of timestamps), and $p_n$ denotes the spatial dimension with a dimensionality equal to the number of locations.

\para{Definition 3 (Mobility Trajectory Generation).} \rev{Given a population distribution $P$, the objective of trajectory generation is to learn a conditional distribution:
\begin{equation}
p_\theta(R \mid P) \approx p_{data}(R \mid P),
\end{equation}
where $p_{data}(R \mid P)$ denotes the true distribution of real trajectories under condition $P$, $p_\theta(\cdot)$ denotes the generation distribution from generator $G$.}

\subsection{Denoising Diffusion Probabilistic Model}
Latent variable models of the type $p_\theta(x_0):=\int p_\theta(x_{0:T}) dx_{1:T}$ are classified as diffusion models. The latents $x_1,..., x_T$ have the same dimensionality as the data $x_0 \sim q(x_0)$.
Two Markov chains are used in the diffusion probabilistic model: a forward chain that turns data into noise and a reverse chain that turns noise back into data. The following Markov chain defines the diffusion process:
\begin{equation}\label{equ:DDPM1}
q\left(\mathbf{x}_{1: T} \mid \mathbf{x}_{0}\right):=\prod_{t=1}^{T} q\left(\mathbf{x}_{t} \mid \mathbf{x}_{t-1}\right) 
\end{equation}
where $q\left(\mathbf{x}_{t} \mid \mathbf{x}_{t-1}\right):=\mathcal{N}\left(\sqrt{1-\beta_{t}} \mathbf{x}_{t-1}, \beta_{t} \mathbf{I}\right)$ and the noise level is denoted by the tiny positive constant $\beta_t$.  Equivalently, $x_t$ can be represented as $x_{t}=\sqrt{\alpha_{t}} x_{0}+\left(1-\alpha_{t}\right) \epsilon$ for $\epsilon \sim \mathcal{N}(0, \mathbf{I})$, where $\alpha_{t}=\sum_{i=1}^{t}\left(1-\beta_{t}\right)$. 

On the other hand, the reverse process, which is described by the following Markov chain, denoises $x_t$ to retrieve $x_0$: 
\begin{equation}\label{equ:DDPM2}
p_{\theta}\left(\mathbf{x}_{0: T}\right):=p\left(\mathbf{x}_{T}\right) \prod_{t=1}^{T} p_{\theta}\left(\mathbf{x}_{t-1} \mid \mathbf{x}_{t}\right)
\end{equation}
where $\mathbf{x}_{T} \sim \mathcal{N}(\mathbf{0}, \mathbf{I})$. Further, $p_\theta(x_{t-1}|x_t)$ is considered to be a normal distribution using learnable parameters as follows:
\begin{equation}\label{equ:DDPM3}
p_{\theta}\left(\mathbf{x}_{t-1} \mid \mathbf{x}_{t}\right):=\mathcal{N}\left(\mathbf{x}_{t-1} ; \boldsymbol{\mu}_{\theta}\left(\mathbf{x}_{t}, t\right), \sigma_{\theta}\left(\mathbf{x}_{t}, t\right) \mathbf{I}\right)
\end{equation}
Ho et al.~\cite{ho2020denoising} has recently proposed denoising diffusion probabilistic models (DDPM), which considers the following specific parameterization of $p_\theta(x_{t-1}|x_t)$:
\begin{equation}\label{equ:DDPM4}
\begin{cases}
\boldsymbol{\mu}_{\theta}\left(\mathbf{x}_{t}, t\right)
= \frac{1}{\alpha_{t}}
  \left(\mathbf{x}_{t}
  - \frac{\beta_{t}}{\sqrt{1-\alpha_{t}}}
    \boldsymbol{\epsilon}_{\theta}\left(\mathbf{x}_{t}, t\right)\right), \\[6pt]
\sigma_{\theta}\left(\mathbf{x}_{t}, t\right)
= \tilde{\beta}_{t}^{1 / 2}
\text{ where }
\tilde{\beta}_{t}=
\begin{cases}
\dfrac{1-\alpha_{t-1}}{1-\alpha_{t}}\,\beta_{t}, & t>1,\\[4pt]
\beta_{1}, & t=1,
\end{cases}
\end{cases}
\end{equation}
where $\epsilon_\theta$ is a trainable denoising function. We denote $\sigma_{\theta}\left(\mathbf{x}_{t}, t\right)$ and $\boldsymbol{\mu}_{\theta}\left(\mathbf{x}_{t}, t\right)$ in Eq.~\ref{equ:DDPM4} as $\sigma^{\mathrm{DDPM}}\left(\mathbf{x}_{t}, t\right)$ and $\mu^{\mathrm{DDPM}}\left(\mathbf{x}_{t}, t, \boldsymbol{\epsilon}_{\theta}\left(\mathbf{x}_{t}, t\right)\right)$ , respectively. According to Ho et al.~\cite{ho2020denoising}, the following goal may be used to train the reverse process:
\begin{equation}\label{equ:DDPM5}
\min _{\theta} \mathcal{L}(\theta):=\min _{\theta} \mathbb{E}_{\mathbf{x}_{0} \sim q\left(\mathbf{x}_{0}\right), \boldsymbol{\epsilon} \sim \mathcal{N}(\mathbf{0}, \mathbf{I}), t}\left\|\boldsymbol{\epsilon}-\boldsymbol{\epsilon}_{\theta}\left(\mathbf{x}_{t}, t\right)\right\|_{2}^{2}, 
\end{equation}
where $\mathbf{x}_{t}=\sqrt{\alpha_{t}} \mathbf{x}_{0}+\left(1-\alpha_{t}\right) \boldsymbol{\epsilon}$. The noise vector contributed to its noisy input $x_t$ is estimated by the denoising function $\epsilon_\theta$.This goal lessens the significance of the lower terms at small $t$ (i.e., low noise levels) and can be thought of as a weighted variational bound of the negative log-likelihood.

\begin{table}[h]
\centering
\caption{\rev{A List of Commonly Used Notations.}}
\rev{\begin{tabular}{ll}
\hline
\textbf{Symbol} & \textbf{Description} \\
\hline
$\mathbf{R} = \{r_1, \dots, r_N\}$ & Mobility trajectory of user $u$ \\
$r_i = (l_i, t_i)$ & $i$-th visit's location and time \\
$\mathbf{P} = \{p_1, \dots, p_N\}$ & Population distribution \\
$\mathcal{G} = (\mathcal{V}, \mathcal{E})$ & Spatial graph, $\mathcal{V}$ is set of location nodes, $\mathcal{E}$ is set of edges \\
$w_{uv}$ & Edge weight between locations $u$ and $v$ (Euclidean distance) \\
$e_l \in \mathbb{R}^d$ & Embedding vector of location $l$ \\
$E_l \in \mathbb{R}^{(|L|+1) \times d}$ & Embedding matrix for all locations \\
$\mathbf{e}_n^t$ & Visited location embedding for time slot $n$ at diffusion step $t$ \\
$t_{\text{emb}}$ & Diffusion step $t$ embedding (positional encoding) \\
$\alpha_t^{(h)}$ & Attention weight of the $h$-th head in multi-head attention \\
$g_\sigma(\cdot)$ & Decoder function \\
$\mathbf{d}$ & Location probability distribution \\
$L_{\text{ind}}$ & Individual mobility-aware loss \\
$L_{\text{pop}}$ & Population distribution-aware loss\\
$\lambda$ & Weight of $L_{\text{pop}}$ \\
\hline
\end{tabular}}
\end{table}

%% file: methods.tex
\section{Methods}
\begin{figure*}[t]
\centering
\includegraphics[width=1.0\textwidth]{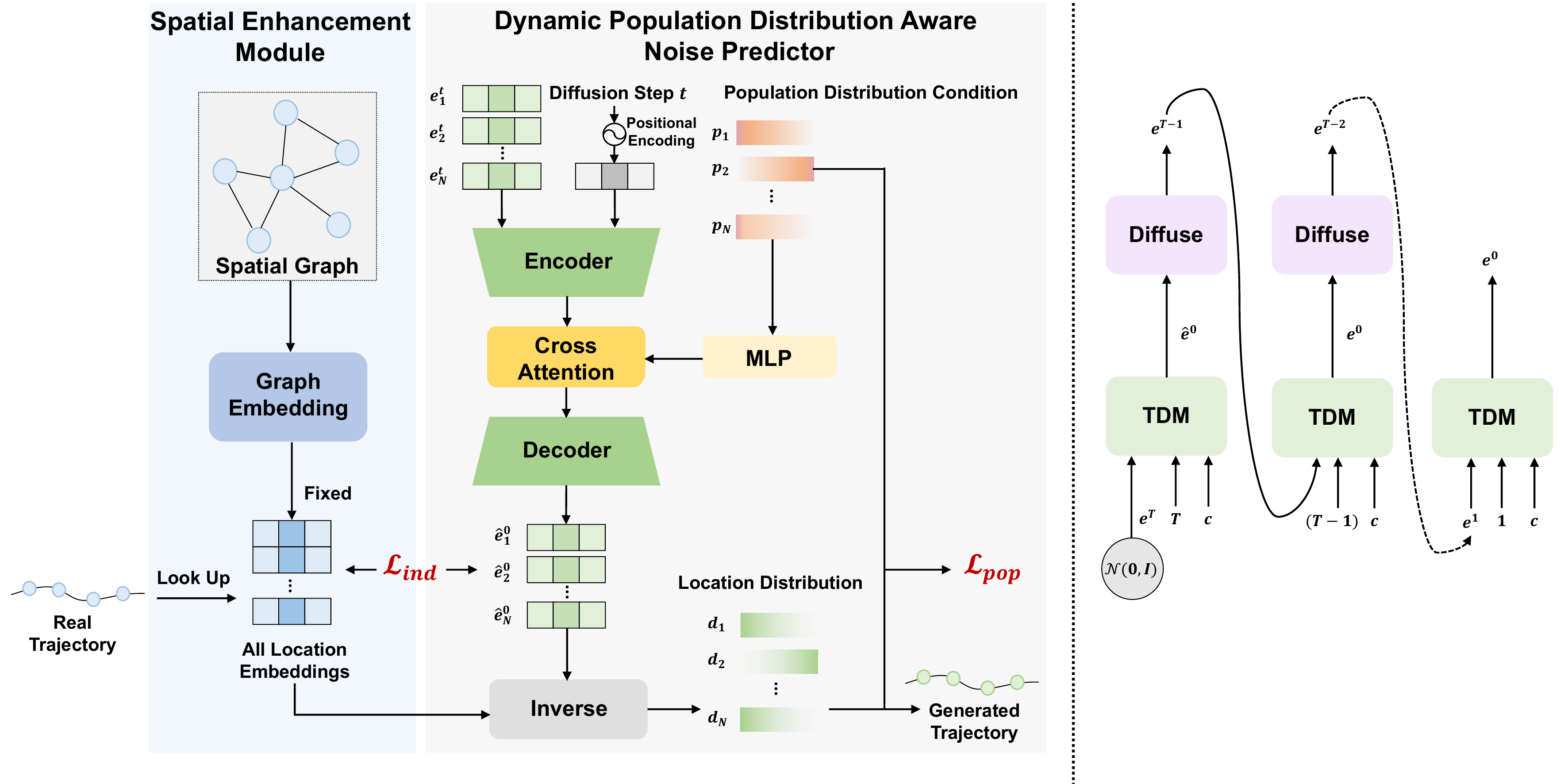}
\caption{(Left) The overview of the Trajectory Diffusion Model (TDM). The model is given a noisy trajectory embedding $e^{t}_{1:N}$ of length $N$ in a diffusion step $t$, as well as the diffusion embedding of diffusion step $t$ and population distribution condition $p_{1:N}$. The denoising network predicts the ultimate clean trajectory embedding $\hat{e}^{t}_{1:N}$ in each sampling step. (Right) Sampling TDM. We sample random noise $e^T$ given a population distribution condition $p_{1:N}$, and then iterate from $T$ to 1. TDM predicts the clean sample $\hat{e}^{0}$ at each step $t$ and diffuses it back to $e^{t-1}$.}
\label{fig:framework}
\end{figure*}

Figure~\ref{fig:framework} illustrates our proposed model, which consists of two components, namely 1) spatial enhancement module and 2) population distribution aware diffusion and denoising processes. Specifically, we design a spatial information enhancement module to model the geographical proximity between locations. Moreover, we develop a dynamic population distribution aware diffusion and denoising module, which enables the accurate generation of trajectories that align with the population distribution.

\subsection{Spatial Enhancement Module}
To solve challenge one, we construct \textbf{spatial enhancement module} and use suitable graph embedding methods to enhance spatial information in the mobility trajectory for choosing the possible next location.
\rev{Classical human mobility models, such as the gravity model~\cite{barbosa2018human} and the intervening opportunities model~\cite{simini2012universal}, explicitly highlight the strong correlation between population movement and geographic distance. Inspired by this, we construct a spatial graph that encodes spatial proximity (e.g., Euclidean distance) as structural information between nodes, thereby introducing an important spatial prior that reflects group-level mobility patterns.
Although the spatial enhancement module does not directly model the dynamics of population distribution, it serves as a structural prior of group mobility. This enables the model to respond more reasonably to spatial structure when integrating population-level constraints.} 

\subsubsection{Graph Construction}
Human mobility exhibits spatial continuity, wherein individuals prefer nearby locations for various activities. For instance, when shopping for groceries, people generally opt for a supermarket close to their residence rather than traveling a distance of 50 km. To enhance the spatial continuity of mobility, we propose the construction of a spatial graph denoted as $G = (V, E)$. This graph captures the locations individuals visit, ensuring a comprehensive representation of their mobility patterns. In the constructed spatial graph, $V$ represents the set of nodes corresponding to all the visited locations. \rev{We model the grids as nodes on the graph. Specifically, the geographical area is divided into grids of different scales. In this work, we set the grid size to 64×64, resulting in 1024 regions. The Geolife dataset has a spatial resolution of approximately 1 km, while the MME dataset has a spatial resolution of approximately 5 km.}
Meanwhile, $E$ represents the set of edges that establish relationships between pairs of locations. Each edge, denoted as $e$, consists of an unordered pair $(u, v)$ and is associated with a positive weight, $w_{uv} > 0$, indicating the Euclidean distance between location $u$ and location $v$. The weight, $w_{uv}$, essentially represents the spatial similarity between the two locations. By employing this approach, we effectively model the spatial continuity of human mobility, enabling a more accurate representation using the spatial graph $G$.

\subsubsection{Spatial Enhancement with Graph Embedding}
To incorporate spatial information into the graph embedding module, we consider the edge values, denoted as $w_{uv}$, in the spatial graph as indicators of the distance between locations $u$ and $v$. This distance value reflects the level of spatial continuity between the two locations. However, a challenge in constructing spatial graphs is sparsity, particularly in edges representing spatial continuity. This sparsity issue is observed when individuals predominantly move within neighboring regions, which is a common characteristic of mobility trajectories. To overcome this limitation and prevent individuals from being confined to a small area, it is crucial to consider first-order proximity (immediate neighbors) and second-order proximity (distant but possible locations) in the fusion of spatial information. By incorporating these perspectives, we ensure that attention is given to nearby and potentially distant locations that may still be relevant to individuals' mobility patterns. This fusion of interactive information between locations enhances the comprehensiveness of the spatial graph, enabling a more accurate representation of human mobility and addressing the sparsity issue associated with spatial continuity edges.

To perform embedding while preserving the graph network topology, we concentrate on the local pairwise proximity between nodes. This local graph network structure comprises the first-order and second-order proximity defined in the spatial graph $G$.

\textbf{1) First-order Proximity.} In the spatial graph $G$, the first-order proximity between two connected nodes $(u, v)$ is represented by the weight $w_{uv}$ associated with their edge. This weight captures the degree of spatial proximity for the linked locations, providing direct information about spatial proximity. However, relying solely on first-order proximity may lead to sparsity issues due to the herd effect in spatial modeling. Therefore, additional measures are required to address this sparse problem.

\textbf{2) Second-order Proximity.} 
To address the sparsity problem, we introduce the concept of second-order proximity. The second-order proximity between nodes $(u, v)$ considers the similarity between their neighborhood network structures. This is computed by comparing the first-order proximities of $u$ and $v$, denoted as $b_u = (w_{u,1}, \dots, w_{u,|V|})$. By incorporating second-order proximity, we can give more attention to remote locations (nodes with smaller edge weights) and effectively address the sparsity problem.

\textbf{3) Spatial Enhancement.} 
To further improve the embedding, we incorporate spatial enhancement information into location representations using graph embedding techniques. By leveraging the constructed spatial graph $G = (V, E)$, the objective is to represent each vertex $v$ in a low-dimensional space $R^d$. We achieve this by learning a function $f_G: V \rightarrow R^d$, where $d << |V|$.

To fuse spatial information from both the first-order and second-order proximity perspectives, we adopt the LINE (Large-scale Information Network Embedding) method~\cite{tang2015line}, a well-known graph embedding approach for large-scale networks. Specifically, we minimize the difference between predicted and actual edge weights for first-order proximity to align with the spatial information. For second-order proximity, random walk and negative sampling techniques are employed among the second-order proximity neighborhoods to mitigate the sparsity issue and avoid the herd effect in trajectories. The optimization process involves separate objective functions for first-order proximities and second-order proximities. Finally, the fixed embedding of all locations is obtained by combining the output embeddings from both first-order and second-order proximity perspectives.
With the spatial graph $G$, the graph embedding with LINE can be denoted as follows:
\begin{equation}\label{equ:LINE}
e_{l}={\rm LINE}_\theta(G),
\end{equation}
An embedding vector $e_l \in \mathbb{R}^d$ is established for each location $l \in L$. A matrix $E_l \in \mathbb{R}^{|L+1| \times d}$ is used to represent all location embeddings.

In summary, we first construct a spatial graph containing all locations that people have visited and then perform graph embedding with LINE from first-order and second-order proximity perspectives to fuse information about spatial continuity into the location embedding.

\subsection{Dynamic Population Distribution Aware Diffusion and Denoising Processes}
To address challenge two and generate individual trajectories that satisfy the population distribution, we design \textbf{dynamic population distribution aware diffusion and denoising processes}. We represent the learning of the spatiotemporal joint distribution of trajectories as a denoising diffusion process, utilizing the refined modeling capabilities of the diffusion model to gradually integrate population distribution as a condition. Furthermore, we design a population distribution aware loss function to ensure that the probability of individuals choosing visitation locations aligns with the population distribution at those geographical locations.

\subsubsection{Diffusion and Denoising Processes}
The trajectory embedding $e$ is the embedding of all locations visited in a day $\{e_1,e_2,...,e_N\}$ of the splicing. Our goal is to learn a distribution model of trajectories. The learning of this distribution is based on a diffusion model~\cite{ho2020denoising}, where each complete trajectory is diffused and denoised. Specifically, we characterize the diffusion process as a Markov process in the space-time domain as $(e^0, e^1,..., e^T)$ for each trajectory sequence with embedding $e$, where $T$ is the number of diffusion steps. We progressively introduce some Gaussian noise into the trajectory embedding from $e^0$ to $e^T$ until they are degraded to pure Gaussian noise. Adding noise is comparable to the process of creating a picture scene, where each pixel receives an individual application of noise~\cite{ho2020denoising}. We diffuse in the spatiotemporal domain according to the following probabilities, respectively:
\begin{equation}\label{equ:ddpm}
q\left(e^{t} \mid e^{t-1}\right):=\mathcal{N}\left(e^{t} ; \sqrt{1-\beta_{t}} e^{t-1}, \beta_{t} \boldsymbol{I}\right),
\end{equation}
where $\beta_{t}$ is the variance schedule for adjusting the intensity of the added noise.

The reconstruction of point $e^n$ is instead represented as a reverse denoising iteration from ${e}^T$ to ${e}^0$. The following is the formulation for the whole reverse denoising process:
\begin{equation}
p_{\theta}\left({e}^{0:T} \right):=p\left({e}^{T}\right) \prod_{t=1}^{T} p_{\theta}\left({e}^{t-1} \mid {e}^{t}\right).
\end{equation}

\subsubsection{Dynamic Population Distribution Aware Noise Predictor}
The inputs to our population distribution aware noise predictor are the diffusion step $t$, the trajectory embedding $e_n$, and the population distribution $p_n$ as the condition. For the diffusion step $t$, we use positional encoding to obtain a 128-dimensional embedding following the previous works~\cite{vaswani2017attention, kong2020diffwave}:
\begin{equation}\label{equ:diffusion step}
\begin{aligned}
t_{\text {emb}}= & {\left[\sin \left(10^{0 \times 4 / 63} t\right), \ldots, \sin \left(10^{63 \times 4 / 63} t\right),\right.} \\
& \left.\cos \left(10^{0 \times 4 / 63} t\right), \ldots, \cos \left(10^{63 \times 4 / 63} t\right)\right].
\end{aligned}
\end{equation}

Our noise predictor consists of the encoder, cross-attention, and decoder.

\textbf{1) Encoder.} To obtain the time dependence of the trajectory sequence, we use an attention mechanism. We introduce a temporal transformer layer to learn temporal correlation as follows:
\begin{equation}
\check{e}^{t}_{n} ={f_{\theta}(e^{t}_{n})},
\end{equation}
where $e^{t}_{n}$ is the visited location embedding for time slot $n$. We utilize transformer~\cite{vaswani2017attention} as the encoder function $f_\theta(\cdot)$ and obtain the encoded vector $\check{e}^{t}_{n}$.

\textbf{2) Cross Attention.} To make the generated individuals satisfy the population distribution as much as possible, we need to add the population distribution to the noise predictor so that the population distribution can influence each generation step. We first map the population distribution $p_n$ to the same dimension as the trajectory embedding $\check{e}_n^t$ as follows:
\begin{equation}\label{equ:attention1}
\check{p_n} = {\rm MLP}_\lambda(p_n).
\end{equation}

To provide more information to the network and stabilize the learning process, we utilize multi-head attention~\cite{vaswani2017attention}. Here, we compute the embeddings by implementing $H$ independent heads as well. In particular, we define the similarity between $\check{e}_n^t$ and $\check{p_n}$, or the similarity between the $n$-th time slot of the trajectory and the $n$-th time slot of the population distribution, as follows:

\begin{equation}\label{equ:attention1}
\alpha_{t}^{(h)}=\frac{\exp \left(\phi^{(h)}\left(\check{e}_n^t, \check{p_n}\right)\right)}{\sum_{g=1}^{N} \exp \left(\phi^{(h)}\left(\check{e}_n^t, \check{p_n}\right)\right)},
\end{equation}

\begin{equation}\label{equ:attention2}
\phi^{(h)}\left(\check{e}_n^t, \check{p_n}\right)=\left\langle W_{Q}^{1(h)} \check{e}_n^t, W_{K}^{1(h)} \check{p_n}\right\rangle,
\end{equation}
where $W_{Q}^{1(h)}, W_{K}^{1(h)} \in \mathbb{R}^{d^{\prime} \times d}$ are transformation matrices and $<,>$ is the inner product function. 

The information of the dynamic population distribution is then aggregated by the similarity $\alpha_{t}^{(h)}$ to provide the trajectory's new embedding vector $\bar{e}_n^t$ for the $t$-th time slot.
\begin{equation}\label{equ:attention3}
\bar{e}_n^{t(h)}=\sum_{k=1}^{T} \alpha_{n, k}^{(h)}\left(W_{V}^{1(h)} \check{e}_n^t\right),
\end{equation}

\begin{equation}\label{equ:attention4}
\bar{e}_n^{t}=\bar{e}_n^{t(1)}\|\bar{e}_n^{t(2)}\| \cdots \| \bar{e}_n^{t(H)},
\end{equation}
where $W_{V}^{1(h)} \in \mathbb{R}^{d^{\prime} \times d}$ is also a transformation matrix, $\|$ is the concatenation operator, and $H$ is the number of total heads.

\textbf{3) Decoder.} The purpose of the decoder is to decode the trajectory embedding $\bar{e}_n^{t}$ into a trajectory embedding $\hat{e}_n^{t}$.
\begin{equation}
\hat{e}_n^{t} ={g_{\sigma}(\bar{e}_n^{t})},
\end{equation}
where the decoder function $g_\sigma(\cdot)$ consists of a convolutional network and a gated activation unit.

In order to transform the real-valued embedding of $\hat{e}_n^{0}$ to the probability distribution of locations $d=\{d_1,d_2,...,d_N\}$ for the discrete space domain, we add an inverse step at the conclusion of the inverse procedure:
\begin{equation}
P(M) = (M^T M)^{-1} M^T,
\end{equation}
where $M$ denotes Embedded Matrix, $M^T$ denotes the transpose of matrix $M$, and $(\cdot)^{-1}$ represents the matrix inverse.

\begin{equation}
d = \hat{e}_n^{0}*P(M),
\end{equation}
where $d$ denotes the probability of visiting each location in the trajectory.

\begin{equation}
i = argmax(d),
\end{equation}
where $argmax(\cdot)$ function is used to find the maximum value index in a vector, and $i$ denotes the location id.

In summary, multiplying the embedding vector by the pseudo-inverse of the embedding matrix using the pseudo-inverse matrix gives the recovered location probability distribution, and then, by finding the index with the highest probability, the recovered ID can be determined. These formulas describe converting the embedding vector to a location ID.

\subsection{Training and Sampling}
The loss function consists of individual mobility aware loss and population distribution aware loss, where individual mobility aware loss is used to encourage the model to restore the mobility characteristics of individuals as much as possible, while population distribution aware loss works by making the probability distribution of visited locations in the generated individual trajectories as similar as possible to the population distribution.

Instead of predicting $\epsilon_t$ as formulated by~\cite{ho2020denoising}, we follow~\cite{ramesh2022hierarchical}and use an equivalent formulation to predict the signal itself, i.e.,
$\hat{e}_n^{0} = \mathcal{F}_\theta(\hat{e}_n^{t}, t)$ with the simple objective~\cite{ho2020denoising} as individual mobility aware loss as follows,
\begin{equation}\label{equ:diffusion}
\mathcal{L}_{\text {ind }}=E_{{e}_n^{0} \sim q\left({e}_n^{0}\right), t \sim[1, T]}\left[\left\|{e}_n^{0}-\mathcal{F}_\theta({\hat{e}_n^{t}, t)}\right\|_{2}^{2}\right].
\end{equation}

We use the cross-entropy loss function as a population distribution aware loss to train the model so that the generated individual trajectories obey the population distribution.
\begin{equation}
\mathcal{L}_{pop}=-\sum_{i=1}^{N} d_{i} \log \left(p_{i}\right),
\end{equation}
where $\mathcal{L}_{pop}$ denotes the cross entropy between distribution $p$ and distribution $d$, and $d_i$ and $p_i$ denote the probabilities of distribution $d$ and distribution $p$ in the $i$-th location, respectively. A smaller cross-entropy indicates a higher similarity between two probability distributions. 

Overall, our training loss is
\begin{equation}\label{equ:location loss}
\mathcal{L}= \mathcal{L}_{ind} + \lambda \mathcal{L}_{pop},
\end{equation}
where $\lambda$ is the weighting factor of population distribution aware loss of $\mathcal{L}_{pop}$.

\subsubsection{Sampling}
Based on ~\cite{ho2020denoising}, $p({{e}_n^{0}})$ is performed iteratively. At each time step $t$, we forecast the clean sample $\hat{e}_n^{0} = \mathcal{F}_\theta(\hat{e}_n^{t}, t)$ and noise it back to ${e}_n^{t-1}$. The process is continued from $t = T$ until ${e}_n^{0}$ is reached.

%% file: experiments.tex
\section{Experiments}

\subsection{Dataset}
\begin{table}[t]
\caption{Basic statistics of mobility datasets.}
\vspace{-10px}
\label{table:datasets}
\begin{center}
\begin{tabular}{ >{\centering\arraybackslash}m{1cm} 
>{\centering\arraybackslash}m{1cm} 
>{\centering\arraybackslash}m{1.2cm} 
>{\centering\arraybackslash}m{1cm}
>{\centering\arraybackslash}m{1cm}
>{\centering\arraybackslash}m{1cm}}
 \hline
 Dataset &City  &Duration &\#Users &\#Loc &\#Traj\\ 
 \hline
 Geolife & Beijing & 5 years & 40 & 3439 & 896\\ 
 MME & Nanchang & 7 days & 6218 & 4096 & 43967\\ 
\hline
\end{tabular}
\end{center}
\end{table}

We performed extensive evaluations on the MME and Geolife datasets, two real-world mobility datasets. \rev{The dynamic population distribution used in our experiments comes from the same data source as the trajectory data and is provided directly by the data vendor after appropriate anonymization and privacy-preserving processing.}
\begin{itemize}
\item \textbf{Geolife}~\cite{zheng2010geolife}: This dataset is collected by Microsoft Research Asia as part of the Geolife project. The dataset encompasses the movement paths of 182 people from April 2007 to August 2012. The mobility trajectories are depicted by a sequence of time-stamped GPS coordinates, each comprising latitude, longitude, and altitude data.
\item \textbf{MME}: The MME dataset is supplied by the China Mobile Research Institute. The mobility data comprises about 6,000 subscriber volumes, mostly focused on Nanchang, covering a temporal range of one week from May 18 to May 24, 2022. Every mobility record in the MME collection comprises an anonymous user ID, a timestamp, and a cellular base station.
\end{itemize}

\textbf{Pre-processing}: In order to exclude users who produced fewer than 10 recordings per day, we preprocessed the trajectory data from both datasets. After ~\cite{chen2019complete}, we change the trajectory to the equal interval trajectory by setting the time interval to 30 minutes. Then, using a granularity of 1 km $\times$ 1 km, we preprocessed the locations to map GPS points to predetermined grid IDs. Table~\ref{table:datasets} summarizes the final comprehensive statistics of two mobility datasets.

\subsection{Baselines}
We compare the performance of our model with six state-of-the-art baselines.

\begin{itemize}
\item \textbf{TimeGEO}~\cite{jiang2016timegeo}: The explore and preferential return (EPR) model~\cite{song2010modelling} is used to model the spatial choices in TimeGEO, a model-based trajectory synthesis approach, whereas dwell rate, burst rate, and the weekly home-based tour number are used to model the temporal options.
\item \textbf{Semi-Markov}~\cite{2016Social}: The exponential distribution in the Semi-Markov process is used to simulate the dwell duration.  To apply a Bayesian inference, the transition matrix and dwell time intensity are modeled using the Dirichlet and gamma priors.
\item \textbf{Hawkes}~\cite{Estimation2016Emmanuel}: The Hawkes process is a popular classical temporal point process in which the intensity function of subsequent points is influenced by the data point that happened.
\item \textbf{LSTM}~\cite{abu2018will}: This model utilizes an LSTM network to forecast the subsequent location and time, thereby treating the predicted results as a generated trajectory.
\revmajor{\item \textbf{GAN}~\cite{ouyang2018non}: The data points in this space are generated using generative adversarial networks, and they will subsequently be converted back to a sequential location trajectory form. }
\revmajor{\item \textbf{TrajGAIL}~\cite{choi2021trajgail}: TrajGAIL is a partially observable Markov decision process that formulates the learning of location sequences in observed trajectories as an imitation learning issue. }
\item \textbf{MoveSim}~\cite{feng2020learning}: The domain knowledge of human movement regularities is included into this generative adversarial framework.
\item \textbf{PateGail}~\cite{wang2023pategail}: It simulates human decision-making by use of the strong generative adversary imitation learning model.
\item \rev{\textbf{TransMob}~\cite{he2020human}: This work generates mobility trajectory data by transferring travel patterns from existing cities.}
\item \rev{\textbf{GTG}~\cite{wang2025gtg}: It leverages universal mobility patterns to capture human mobility preferences for trajectory generation.}
\end{itemize}

\subsection{Experimental Settings}
\subsubsection{Metrics}
We evaluate the extent and effectiveness of the generated trajectories in retaining the real trajectories' statistical characteristics. The eight evaluation metrics we chose consider measures of the statistical characteristics of the mobility trajectories and measures of some macroscopic distributions.

The measurement of the statistical characteristics of the mobility trajectories considers the measurement of the temporal and spatial statistical characteristics. It is described in detail in the following:
\begin{itemize}
\item \textbf{Distance}: This is a spatial statistical metric used to calculate the distance between neighboring mobility data in a trajectory.
\item \textbf{Radius}: The root mean square of the distance between each location point on a single trajectory and its center of mass is known as the radius of gyration.
\item \textbf{Duration}: This metric quantifies the temporal statistical attributes to assess the time users spend in various locations.
\item \textbf{DailyLoc}: Daily visited locations are determined by how many locations each user visits each day.
\item \textbf{G-rank}: The top visited frequency to various areas for all users is measured by this metric of geographical statistical attributes.
\item \textbf{I-rank}: An individual version of G-rank.
\end{itemize}

Measurements of some macroscopic distributions were considered for the population distribution as well as for the OD matrix and are described in detail in the following:
\begin{itemize}
\item \textbf{Population Distribution}: This metric measures population distribution in all regions.
\item \textbf{OD Similarity}: This metric measures the OD matrix formed by aggregating individual trajectories.
\end{itemize}

In particular, probability distributions are used to represent the metrics of Population Distribution, G-rank, I-rank, Distance, Radius, Duration, and DailyLoc. We utilize the Jensen-Shannon scatter (JSD) to quantify the differences between the created sequences and the genuine sequences in order to visually assess how comparable they are. Specifically, the JSD between two distributions $\boldsymbol{p}$ and $\boldsymbol{q}$ may be defined as follows:
\begin{equation}\label{equ:JSD}
{\rm JSD}(\boldsymbol{p},\boldsymbol{q})=\frac{1}{2}{\rm KL}(\boldsymbol{p}||\frac{\boldsymbol{p}+\boldsymbol{q}}{2})+\frac{1}{2}{\rm KL}(\boldsymbol{q}||\frac{\boldsymbol{p}+\boldsymbol{q}}{2}),
\end{equation}
where the Kullback-Leibler divergence is denoted by ${\rm KL}(\cdot||\cdot)$. In contrast, the OD matrix's metric is not a probability distribution function. Consequently, we employ the cosine similarity to assess the similarity between the OD matrix aggregated from real trajectories and the OD matrix aggregated from generated trajectories.

\begin{table}[t]
	\centering
	\caption{\revmajor{The details of hyper-parameters settings.}}
	\setlength\tabcolsep{4pt}
	\scalebox{0.9}{
	\begin{tabular}{l|c}
		\toprule[1pt]
		\textbf{\revmajor{Hyper-parameters}} & \textbf{\revmajor{Settings}}\\ \hline
		\revmajor{The embedding size of diffusion step} & \revmajor{128} \\ \hline
		\revmajor{The embedding size of location embedding} & \revmajor{128} \\ \hline
        \revmajor{The channel number of the convolutional network} & \revmajor{64} \\ \hline
        \revmajor{Transformer layer number} & \revmajor{4} \\ \hline
        \revmajor{Transformer head number} & \revmajor{8} \\ \hline
        \revmajor{Learning rate} & \revmajor{1e-3} \\ \hline
        \revmajor{Diffusion steps} & \revmajor{1000} \\ \hline
        \revmajor{The weighting factor of population distribution} & \revmajor{0.5} \\ \hline
		\revmajor{Batch size} & \revmajor{16}\\

\bottomrule[1pt]
	\end{tabular}}
	\label{tab::hyper-parameters}
	\vspace{-0.2cm}
\end{table}
\subsubsection{Hyper-parameters Settings}
According to the user, the first 70\% of the dataset is used for training, the second 10\% is used for validation, and the last 20\% is used for testing. In terms of our model's hyperparameter settings, we set the diffusion step and location embedding sizes to 128 and the convolutional network's channels to 64, transformer layers to 4, and attention mechanism heads to 8. Then, during the training phase, we set the batch size to be 16, the diffusion steps to be 1000, and the learning rate to be 1e-3. Table~\ref{tab::hyper-parameters} displays the specifics of the hyper-parameters settings. Lastly, Pytorch is used to implement the suggested framework. A Linux server with eight GPUs (NVIDIA RTX 2080 Ti * 8) is used for training. In reality, two GPUs can efficiently train our framework in 12 hours.

\subsection{Overall Performance}
\begin{table*}[ht]
\center
\caption{Performance comparisons on the GeoLife dataset, where bold denotes best (lowest) results and underline denotes the second best results. Pop. Dis. denotes Population Distribution, and OD Sim. denotes Origin–Destination Similarity.}
\label{tab::performance_geolife}
\small
\setlength{\tabcolsep}{0.8mm}{\begin{tabular}{l|cccccccc}
\hline
\multirow{2}{*}{Algorithm}
    & \textbf{Distance}
    & \textbf{Radius}  
    & \textbf{Duration}
    & \textbf{Daily-Loc}
    & \textbf{G-rank}  
    &\textbf{I-rank}
    & \textbf{Pop. Dis.}  
    &\textbf{OD Sim.}
    \\
    & (JSD)   
    & (JSD)   
    & (JSD)   
    & (JSD)   
    & (JSD)   
    & (JSD)   
    & (JSD)   
    & (Cosine Similarity)   \\\hline
Semi-Markov
    & 0.2028	
    & 0.5276	
    & 0.1142	
    & 0.6513	
    & 0.1285	
    & 0.0835	
    & 0.6219	
    & 0.1017
    \\
Hawkes
    & 0.2371	
    & 0.5053	
    & 0.1398	
    & 0.6272	
    & 0.0963	
    & 0.0935	
    & 0.5384	
    & 0.1165
   \\
LSTM
    & 0.0473	
    & 0.0708	
    & 0.0935	
    & 0.4867	
    & 0.0578	
    & 0.5864	
    & 0.4435	
    & 0.1248
  \\
TimeGeo
    & 0.0458	
    & 0.0712	
    & 0.0826	
    & 0.5474	
    & 0.0531	
    & 0.0592	
    & 0.3576	
    & 0.3543
    \\
\revmajor{GAN}
    & \revmajor{0.0272}
    & \revmajor{0.0919}	
    & \revmajor{0.0701}
    & \revmajor{0.3563}	
    & \revmajor{0.0547}	
    & \revmajor{0.0435}	
    & \revmajor{0.4290}	
    & \revmajor{0.1964}
    \\
\revmajor{TrajGAIL}
    & \revmajor{0.0204}	
    & \revmajor{0.0853}	
    & \revmajor{0.0746}
    & \revmajor{0.3015}	
    & \revmajor{0.0556}	
    & \revmajor{0.0491}	
    & \revmajor{0.3762}	
    & \revmajor{0.2148}
    \\
MoveSim
    & \underline{0.0134}	
    & 0.2763	
    & \underline{0.0672}
    & 0.2684	
    & 0.0518	
    & 0.0425	
    & 0.4162	
    & 0.1316
    \\
\rev{TransMob}
    & \rev{0.2314}	
    & \rev{0.8964}
    & \rev{0.0760}	
    & \rev{0.3382}	
    & \rev{0.0553}	
    & \rev{0.0519}	
    & \rev{0.4081}	
    & \rev{0.2306}
    \\
\rev{GTG}
    & \rev{0.1681}	
    & \rev{0.0805}
    & \rev{0.0726}	
    & \rev{0.2472}	
    & \rev{0.4693}	
    & \rev{0.0385}	
    & \rev{0.0361}	
    & \rev{0.3240}
    \\
PateGail
    & 0.0152	
    & \underline{0.0691}
    & 0.0804	
    & \underline{0.2217}	
    & \textbf{0.0305}	
    & \underline{0.0317}	
    & \underline{0.3039}	
    & \underline{0.3819}
    \\
Ours
    & \textbf{0.0085}	
    & \textbf{0.0588}	
    & \textbf{0.0501}	
    & \textbf{0.1032}	
    & \underline{0.4274}
    & \textbf{0.0296}	
    & \textbf{0.1595}	
    & \textbf{0.7973}          
    \\
Improvement
    & 36.58\%        
    & 14.91\%
    & 25.45\%         
    & 53.45\%         
    & -     
    & 6.62\%  
    & 47.52\%
    & 108.77\%
    \\\hline
\end{tabular}
}
\end{table*}

\begin{table*}[ht]
\center
\caption{Performance comparisons on the MME dataset, where bold denotes best (lowest) results and underline denotes the second best results. Pop. Dis. denotes Population Distribution, and OD Sim. denotes Origin–Destination Similarity.}
\label{tab::performance_mme}
\small
\setlength{\tabcolsep}{0.8mm}{\begin{tabular}{l|cccccccc}
\hline
\multirow{2}{*}{Algorithm}
    & \textbf{Distance}
    & \textbf{Radius}  
    & \textbf{Duration}
    & \textbf{Daily-Loc}
    & \textbf{G-rank}  
    &\textbf{I-rank}
    & \textbf{Pop. Dis.}  
    &\textbf{OD Sim.}
    \\
    & (JSD)   
    & (JSD)   
    & (JSD)   
    & (JSD)   
    & (JSD)   
    & (JSD)   
    & (JSD)   
    & (Cosine Similarity)   \\\hline
Semi-Markov
    & 0.0498	
    & 0.4683	
    & 0.0828	
    & 0.4362	
    & 0.0311	
    & 0.0326	
    & 0.4517	
    & 0.1126
    \\
Hawkes
    & 0.0517	
    & 0.4826	
    & 0.0913	
    & 0.4128	
    & 0.0305	
    & 0.0329	
    & 0.4865	
    & 0.1084
   \\
LSTM
    & 0.0321	
    & 0.3956	
    & 0.0585	
    & 0.2951	
    & 0.0232	
    & 0.0287	
    & 0.3895	
    & 0.1922
  \\
TimeGeo
    & 0.0443	
    & 0.4126	
    & 0.0623	
    & 0.3868	
    & 0.0283	
    & 0.0348	
    & 0.2955	
    & 0.3800
    \\
\revmajor{GAN}
    & \revmajor{0.0231}	
    & \revmajor{0.3602}	
    & \revmajor{0.0685}
    & \revmajor{0.3264}	
    & \revmajor{0.0258}	
    & \revmajor{0.0281}	
    & \revmajor{0.3017}	
    & \revmajor{0.2943}
    \\
\revmajor{TrajGAIL}
    & \revmajor{0.0285}	
    & \revmajor{0.3712}	
    & \revmajor{0.0604}
    & \revmajor{0.3029}	
    & \revmajor{0.0246}	
    & \revmajor{0.0267}	
    & \revmajor{0.3331}	
    & \revmajor{0.2593}
    \\
MoveSim
    & 0.0225	
    & 0.3630	
    & 0.0568	 
    & 0.2635	
    & 0.0225	 
    & 0.0253	
    & 0.3979	
    & 0.1469
    \\
\rev{TransMob}
    & \rev{0.2543}	
    & \rev{0.3870}
    & \rev{0.0541}	
    & \rev{0.3914}	
    & \rev{0.0236}	
    & \rev{0.0278}	
    & \rev{0.3584}	
    & \rev{0.2315}
    \\
\rev{GTG}
    & \rev{0.0107}	
    & \rev{0.2942}
    & \rev{\underline{0.0493}}	
    & \rev{0.2374}	
    & \rev{0.0208}	
    & \rev{0.0230}	
    & \rev{0.2753}	
    & \rev{0.3079}
    \\
PateGail
    & \underline{0.0004}	
    & \underline{0.1817}	
    & 0.0789	
    & \underline{0.0902}	
    & \textbf{0.0129}	
    & \textbf{0.0173}	
    & \underline{0.2406}	
    & \underline{0.3888}
    \\
Ours
    & \textbf{0.0002}	
    & \textbf{0.1084}	
    & \textbf{0.0321}	
    & \textbf{0.0105}	
    & \underline{0.0141}
    & \underline{0.0184}	
    & \textbf{0.0998}	
    & \textbf{0.8396 }         
    \\
Improvement
    & 50.00\%        
    & 40.34\%    
    & 43.49\%          
    & 88.36\%          
    & -     
    & -  
    & 58.52\%
    & 115.95\% 
    \\\hline
\end{tabular}
}
\end{table*}

The performance of our model and the baseline approach are compared on two real datasets in Table~\ref{tab::performance_geolife} and Table~\ref{tab::performance_mme}. In particular, we generate 13,000 trajectories at random using each generation method, and then we compare them with the relevant features of the real trajectories. We use different metrics to measure the mobility trajectories' statistical features and macroscopic distribution, including spatial and temporal statistical features. The performance comparison leads to the following conclusions:
\begin{itemize}
\item Deep generative models outperform non-deep generative models in generating trajectories while preserving the statistical characteristics of real trajectories. This is due to the deep generative models' ability to extract richer and higher-level features from trajectory data, enabling them to capture better the latent patterns and structures in the trajectory data and adapt to the complex distribution and non-linear relationships in the data, resulting in more accurate and realistic trajectory generation.
\item Our model performs best on most metrics, which can be attributed to the diffusion model's unique diffusion and denoising process, which helps to better learn the distribution of trajectories. These results suggest that diffusion models should be implemented in future investigations that involve spatiotemporal data generation.
\item Our model is much better than the other baselines in the measures of macroscopic distribution (population distribution and OD similarity), which proves that our model's unique design of population distribution perception works. The introduction of a loss function for population distribution perception and the inclusion of population distribution information in each step of the denoising process sufficiently allow our diffusion model-based trajectory generation model to perceive the population distribution.
\end{itemize}

\begin{table*}[ht]
\center
\caption{Results of the ablation study regarding different metrics on the GeoLife dataset. Pop. Dis. denotes Population Distribution, and OD Sim. denotes Origin–Destination Similarity.}
\label{tab::ablation1}
\small
\resizebox{\textwidth}{!}{
\setlength{\tabcolsep}{0.8mm}{\begin{tabular}{l|cccccccc}
\hline
\multirow{2}{*}{Algorithm}
    & \textbf{Distance}
    & \textbf{Radius}  
    & \textbf{Duration}
    & \textbf{Daily-Loc}
    & \textbf{G-rank}  
    &\textbf{I-rank}
    & \textbf{Pop. Dis.}  
    &\textbf{OD Sim.}
    \\
    & (JSD) 
    & (JSD)  
    & (JSD)   
    & (JSD)   
    & (JSD)   
    & (JSD)   
    & (JSD)  
    & (Cosine Similarity)   \\\hline
No-Population
    & 0.0095
    & 0.0608	
    & 0.0428	
    & 0.0975	
    & 0.4381	
    & 0.0313
    & 0.2638	
    & 0.4971
    \\
No-Spatial Enhancement
    & 0.0137	
    & 0.0652
    & 0.0457	
    & 0.1006	
    & 0.4532	 
    & 0.0341	
    & 0.1949	
    & 0.6825
   \\
Our model
    & 0.0085	
    & 0.0588	
    & 0.0501
    & 0.1032	
    & 0.4274
    & 0.0296	
    & 0.1595
    & 0.7973  
    \\\hline
\end{tabular}
}
}
\end{table*}

\begin{table*}[ht]
\center
\caption{Results of the ablation study regarding different metrics on the MME dataset. Pop. Dis. denotes Population Distribution, and OD Sim. denotes Origin–Destination Similarity.}
\label{tab::ablation2}
\small
\resizebox{\textwidth}{!}{
\setlength{\tabcolsep}{0.8mm}{\begin{tabular}{l|cccccccc}
\hline
\multirow{2}{*}{Algorithm}
    & \textbf{Distance}
    & \textbf{Radius}  
    & \textbf{Duration}
    & \textbf{Daily-Loc}
    & \textbf{G-rank}  
    &\textbf{I-rank}
    & \textbf{Pop. Dis.}  
    &\textbf{OD Sim.}
    \\
    & (JSD)   
    & (JSD)   
    & (JSD)   
    & (JSD)   
    & (JSD)   
    & (JSD)   
    & (JSD)   
    & (Cosine Similarity)   \\\hline
No-Population
    & 0.0275	
    & 0.3351	
    & \textbf{0.0252}	
    & 0.0248	
    & 0.0479	
    & 0.0242
    & 0.2374	
    & 0.5964
    \\
No-Spatial Enhancement
    & 0.0302	
    & 0.3561	
    & 0.0284	
    & 0.0296	
    & 0.0415	 
    & 0.0227	
    & 0.1839	
    & 0.7301
   \\
Our model
    & \textbf{0.0002}	
    & \textbf{0.1084}	
    & 0.0321	
    & \textbf{0.0105}	
    & \textbf{0.0141}	
    & \textbf{0.0184}	
    & \textbf{0.0998}	
    & \textbf{0.8396}
    \\\hline
\end{tabular}
}
}
\end{table*}

\begin{table*}[ht]
\centering
\caption{\rev{Ablation study on spatial enhancement module on the GeoLife dataset. Pop. Dis. denotes Population Distribution, OD Sim. denotes Origin–Destination Similarity, 
and CL denotes Contrastive Learning.}}
\label{tab::ablation3}
\small
\setlength{\tabcolsep}{0.8mm}
\rev{
\begin{tabular}{c|cccccccc}
\hline
Spatial  & \textbf{Distance} & \textbf{Radius} & \textbf{Duration} & \textbf{Daily-Loc} & \textbf{G-rank} & \textbf{I-rank} & \textbf{Pop. Dis.} & \textbf{OD Sim.} \\
Module   & (JSD)             & (JSD)           & (JSD)             & (JSD)              & (JSD)           & (JSD)           & (JSD)                           & (Cosine Similarity) 
    \\\hline
LINE & \textbf{0.0085} & \textbf{0.0588} & \textbf{0.0501} & \textbf{0.1032} & \underline{0.4274} & \underline{0.0296} & \textbf{0.1595} & \textbf{0.7973} \\
GNN  & 0.0092 & 0.0625 & 0.0538 & 0.1094 & \textbf{0.4180} & \textbf{0.0283} & 0.1731 & 0.7842 \\
CL & 0.0096 & 0.0641 & 0.0545 & 0.1112 & 0.4330 & 0.0314 & 0.1810 & 0.7764
\\\hline
\end{tabular}
}
\end{table*}

\begin{table*}[ht]
\center
\caption{\rev{Ablation study on spatial enhancement module on the MME dataset. Pop. Dis. denotes Population Distribution, OD Sim. denotes Origin–Destination Similarity, 
and CL denotes Contrastive Learning.}}
\label{tab::ablation4}
\small
\setlength{\tabcolsep}{0.8mm}{\rev{\begin{tabular}{c|cccccccc}
\hline
Spatial  & \textbf{Distance} & \textbf{Radius} & \textbf{Duration} & \textbf{Daily-Loc} & \textbf{G-rank} & \textbf{I-rank} & \textbf{Pop. Dis.} & \textbf{OD Sim.} \\
Module   & (JSD)             & (JSD)           & (JSD)             & (JSD)              & (JSD)           & (JSD)           & (JSD)                           & (Cosine Similarity) 
    \\\hline
LINE
& \underline{0.0002} 
& \underline{0.1084}
& \textbf{0.0321}
& \underline{0.0105}
& \underline{0.0141}
& \underline{0.0184}
& \textbf{0.0998}
& \textbf{0.8396} \\
GNN 
& 0.0003 
& 0.1160 
& 0.0348 
& 0.0123 
& 0.0153 
& 0.0201 
& 0.1086 
& 0.8230 \\
CL
& \textbf{0.0002} 
& \textbf{0.1052} 
& \underline{0.0330} 
& \textbf{0.0098} 
& \textbf{0.0137} 
& \textbf{0.0175} 
& \underline{0.1013} 
& \underline{0.8351} 
    \\\hline
\end{tabular}}
}
\end{table*}

\subsection{Ablation Study}
In order to evaluate the significance of the various components discussed in our research, we conducted an ablation study on two mobility datasets. This involved the removal of each component from the complete model. The results were also assessed using eight carefully chosen metrics. 
The results of both datasets are presented in Table~\ref{tab::ablation1} and ~\ref{tab::ablation2}.
In particular, the complete version of our proposed model is represented by \textbf{Our Model}, \textbf{No-Population} represents the design without population distribution perception (including the loss function for population distribution perception and the introduction of population distribution in the denoising network), and \textbf{No-Spatial Enhancement} represents the removal of the spatial enhancement module. We observed that upon removing the spatial enhancement module, spatial metrics such as distance and radius significantly decreased, highlighting the module's effectiveness in capturing the spatial relationships in human mobility behaviors. Similarly, when population distribution information is removed, macro-distribution metrics like population distribution and OD similarity significantly drop, underscoring the importance of incorporating population distribution awareness in generating trajectories that align with the real-world macro distribution. Overall, with the support of these two core designs, our proposed method outperforms all the baseline methods across most of the evaluation metrics.

\rev{Moreover, we conducted ablation studies on the GeoLife and MME datasets by implementing the spatial enhancement module with three alternatives: LINE, GNN, and contrastive learning. LINE learns node embeddings for the spatial graph by modeling both first-order proximity (co-occurrence relationships between directly connected nodes) and second-order proximity (similarity of node neighborhood structures). Once trained, the embeddings can be directly reused in downstream tasks without retraining. In the GNN setting, node representations are learned end-to-end, jointly optimized with downstream task parameters, and require re-initialization and retraining for each new task. The contrastive learning approach is based on the InfoNCE loss, where geographically close node pairs are treated as positive samples and distant pairs as negative samples; the resulting embeddings are also used as fixed spatial priors in the downstream model. As shown in Tables~\ref{tab::ablation3} and~\ref{tab::granularity1}, LINE achieves the best or near-best results on most key metrics, consistently outperforming the others in population distribution and OD similarity, which are closely related to the task objectives. Compared with the other methods, LINE offers stable performance. While GNN and contrastive learning have slight advantages on individual metrics, the overall differences are minor, and LINE demonstrates more balanced and robust performance across both datasets. LINE also provides high efficiency and reusability since it requires only a single pretraining step to provide stable spatial priors across tasks, whereas GNN demands repeated end-to-end training, and contrastive learning is prone to performance drops under imbalanced positive–negative sample distributions, leading to unstable performance in different scenarios. Overall, LINE achieves the best trade-off among performance, efficiency, and reusability and is therefore adopted as the spatial enhancement module in our framework.}

\subsection{\rev{Sensitivity Analysis}}
\subsubsection{\rev{Sensitivity Analysis of Hyperparameters}} 
\rev{We conduct a sensitivity analysis of key hyperparameters, in which hyperparameter $\lambda$ plays an important role in balancing individual-level fidelity and group-level consistency. As shown in Table~\ref{tab::hyperparameters1} and ~\ref{tab::hyperparameters2}, the experimental results show that when this parameter is set to a smaller value, the model focuses more on preserving the fidelity of individual trajectories. In contrast, when it is set to a larger value, the model tends to prioritize maintaining the overall consistency of the group distribution. These results show that $\lambda=0.5$ achieves the most balanced performance: it ensures the smallest degradation in the six individual-level metrics while yielding the largest improvements in the two group-level metrics. Therefore, we set the parameter $\lambda$ to 0.5.
}

\begin{table}[htbp]
\centering
\caption{\rev{Sensitivity analysis of the hyperparameters $\lambda$ on the GeoLife dataset. Pop. Dis. denotes Population Distribution, and OD Sim. denotes Origin–Destination Similarity. }}
\label{tab::hyperparameters1}
\resizebox{\textwidth}{!}{
\rev{
\setlength{\tabcolsep}{0.8mm}{\begin{tabular}{c|cccccccc}
\toprule
\multirow{2}{*}{Weight $\lambda$}
    & \textbf{Distance}
    & \textbf{Radius}  
    & \textbf{Duration}
    & \textbf{Daily-Loc}
    & \textbf{G-rank}  
    & \textbf{I-rank}
    & \textbf{Pop. Dis.}  
    &\textbf{OD Sim.}
    \\
    & (JSD) 
    & (JSD)  
    & (JSD)   
    & (JSD)   
    & (JSD)   
    & (JSD)   
    & (JSD)  
    & (Cosine Similarity)   \\\hline
0.25 & 0.0078 & 0.0563 & 0.0482 & 0.0990 & 0.4213 & 0.0288 & 0.2875 & 0.5821 \\
0.50 & 0.0085 & 0.0588 & 0.0501 & 0.1032 & 0.4274 & 0.0296 & 0.1595 & 0.7973 \\
0.75 & 0.0139 & 0.0772 & 0.0646 & 0.1793 & 0.4697 & 0.0345 & 0.1360 & 0.8417 \\
1.00 & 0.0206 & 0.0855 & 0.0703 & 0.2558 & 0.5290 & 0.4362 & 0.1232 & 0.8645 \\
\bottomrule
\end{tabular}}}}
\end{table}

\begin{table}[htbp]
\centering
\caption{\rev{Sensitivity analysis of the hyperparameters $\lambda$ on the MME dataset. Pop. Dis. denotes Population Distribution, and OD Sim. denotes Origin–Destination Similarity. }}
\label{tab::hyperparameters2}
\resizebox{\textwidth}{!}{
\rev{
\setlength{\tabcolsep}{0.8mm}{\begin{tabular}{c|cccccccc}
\toprule
\multirow{2}{*}{Weight $\lambda$}
    & \textbf{Distance}
    & \textbf{Radius}  
    & \textbf{Duration}
    & \textbf{Daily-Loc}
    & \textbf{G-rank}  
    & \textbf{I-rank}
    & \textbf{Pop. Dis.}  
    &\textbf{OD Sim.}
    \\
    & (JSD) 
    & (JSD)  
    & (JSD)   
    & (JSD)   
    & (JSD)   
    & (JSD)   
    & (JSD)  
    & (Cosine Similarity)   \\\hline
0.25 & 0.0001 & 0.1012 & 0.0307 & 0.0098 & 0.0135 & 0.0178 & 0.1420 & 0.7521 \\
0.50 & 0.0001 & 0.1084 & 0.0321 & 0.0105 & 0.0141 & 0.0184 & 0.0998 & 0.8396 \\
0.75 & 0.0004 & 0.1387 & 0.0396 & 0.0132 & 0.0171 & 0.0209 & 0.0875 & 0.8562 \\
1.00 & 0.0009 & 0.1869 & 0.0470 & 0.0180 & 0.0208 & 0.0242 & 0.0821 & 0.8714 \\
\bottomrule
\end{tabular}}}}
\end{table}

\subsubsection{\rev{Sensitivity Analysis of Spatial Granularity}} 
\rev{As shown in Table~\ref{tab::granularity1} and~\ref{tab::granularity2}, we conduct a sensitivity analysis of the model's performance under different levels of spatial granularity. Specifically, spatial granularity indicates the degree of refinement with which the space is partitioned into grids. The number of grids can be obtained by dividing the city’s bounding box (based on the minimum and maximum latitude and longitude) into equally spaced grids at different resolutions: 32×32, 64×64, and 128×128. The results show that although finer grid granularity, which corresponds to a larger number of grids, makes the generation task more challenging as the set of candidate locations becomes larger, the model’s performance does not degrade significantly, maintaining nearly 80\% stability overall. This demonstrates the robustness of our model, as it consistently maintains strong performance across varying levels of spatial resolution.}

\begin{table}[htbp]
\centering
\caption{\rev{Sensitivity analysis of spatial granularity on the GeoLife dataset. Pop. Dis. denotes Population Distribution, and OD Sim. denotes Origin–Destination Similarity.  }}
\label{tab::granularity1}
\resizebox{\textwidth}{!}{
\rev{
\setlength{\tabcolsep}{0.8mm}{\begin{tabular}{c|cccccccc}
\hline
    Spatial
    & \textbf{Distance}
    & \textbf{Radius}  
    & \textbf{Duration}
    & \textbf{Daily-Loc}
    & \textbf{G-rank}  
    & \textbf{I-rank}
    & \textbf{Pop. Dis.}  
    &\textbf{OD Sim.}
    \\
    Granularity
    & (JSD) 
    & (JSD)  
    & (JSD)   
    & (JSD)   
    & (JSD)   
    & (JSD)   
    & (JSD)  
    & (Cosine Similarity)   \\\hline
32$\times$32  & 0.0073 & 0.0554 & 0.0485 & 0.0991 & 0.4180 & 0.0281 & 0.1467 & 0.8142 \\
64$\times$64  & 0.0085 & 0.0588 & 0.0501 & 0.1032 & 0.4274 & 0.0296 & 0.1595 & 0.7973 \\
128$\times$128 & 0.0107 & 0.0679 & 0.0556 & 0.1204 & 0.4510 & 0.0315 & 0.1764 & 0.7628 \\
\bottomrule
\end{tabular}}}}
\end{table}

\begin{table}[htbp]
\centering
\caption{\rev{Sensitivity analysis of spatial granularity on the MME dataset. Pop. Dis. denotes Population Distribution, and OD Sim. denotes Origin–Destination Similarity.  }}
\label{tab::granularity2}
\resizebox{\textwidth}{!}{
\rev{
\setlength{\tabcolsep}{0.8mm}{\begin{tabular}{c|cccccccc}
\hline
Spatial
    & \textbf{Distance}
    & \textbf{Radius}  
    & \textbf{Duration}
    & \textbf{Daily-Loc}
    & \textbf{G-rank}  
    & \textbf{I-rank}
    & \textbf{Pop. Dis.}  
    &\textbf{OD Sim.}
    \\
Granularity
    & (JSD) 
    & (JSD)  
    & (JSD)   
    & (JSD)   
    & (JSD)   
    & (JSD)   
    & (JSD)  
    & (Cosine Similarity)   \\\hline
32$\times$32  & 0.0001 & 0.1072 & 0.0307 & 0.0098 & 0.0139 & 0.0181 & 0.0983 & 0.8424 \\
64$\times$64  & 0.0001 & 0.1084 & 0.0321 & 0.0105 & 0.0141 & 0.0184 & 0.0998 & 0.8396 \\
128$\times$128 & 0.0003 & 0.1193 & 0.0355 & 0.0124 & 0.0162 & 0.0205 & 0.1226 & 0.7931 \\
\bottomrule
\end{tabular}}}}
\end{table}

\subsubsection{\rev{Sensitivity Analysis of Time Interval}} 
\rev{As shown in Table~\ref{tab::time1} and~\ref{tab::time2}, we conduct a sensitivity analysis of the model's performance under different temporal intervals. Specifically, we set the sampling intervals of the trajectories to 15 minutes, 30 minutes, and 60 minutes. The experimental results show that although finer intervals introduce more stochastic trajectory points per day and increase the complexity of the generation task, the model’s overall performance does not fluctuate significantly. This demonstrates the robustness of the proposed model, which maintains stable and strong performance across different temporal resolutions.
}

\begin{table}[htbp]
\centering
\caption{\rev{Sensitivity analysis of time interval on the GeoLife dataset. Pop. Dis. denotes Population Distribution, and OD Sim. denotes Origin–Destination Similarity.  }}
\label{tab::time1}
\resizebox{\textwidth}{!}{
\rev{
\setlength{\tabcolsep}{0.8mm}{\begin{tabular}{c|cccccccc}
\toprule
\multirow{2}{*}{Interval (min)}
    & \textbf{Distance}
    & \textbf{Radius}  
    & \textbf{Duration}
    & \textbf{Daily-Loc}
    & \textbf{G-rank}  
    & \textbf{I-rank}
    & \textbf{Pop. Dis.}  
    &\textbf{OD Sim.}
    \\
    & (JSD) 
    & (JSD)  
    & (JSD)   
    & (JSD)   
    & (JSD)   
    & (JSD)   
    & (JSD)  
    & (Cosine Similarity)   \\\hline
15   & 0.0097 & 0.0597 & 0.0517 & 0.1055 & 0.4302 & 0.0301 & 0.1624 & 0.7902 \\
30   & 0.0085 & 0.0588 & 0.0501 & 0.1032 & 0.4274 & 0.0296 & 0.1595 & 0.7973 \\
60   & 0.0078 & 0.0584 & 0.0498 & 0.1027 & 0.4261 & 0.0293 & 0.1582 & 0.78011 \\
\bottomrule
\end{tabular}}}}
\end{table}

\begin{table}[htbp]
\centering
\caption{\rev{Sensitivity analysis of time interval on the MME dataset. Pop. Dis. denotes Population Distribution, and OD Sim. denotes Origin–Destination Similarity.  }}
\label{tab::time2}
\resizebox{\textwidth}{!}{
\rev{
\setlength{\tabcolsep}{0.8mm}{\begin{tabular}{c|cccccccc}
\toprule
\multirow{2}{*}{Interval (min)}
    & \textbf{Distance}
    & \textbf{Radius}  
    & \textbf{Duration}
    & \textbf{Daily-Loc}
    & \textbf{G-rank}  
    & \textbf{I-rank}
    & \textbf{Pop. Dis.}  
    &\textbf{OD Sim.}
    \\
    & (JSD) 
    & (JSD)  
    & (JSD)   
    & (JSD)   
    & (JSD)   
    & (JSD)   
    & (JSD)  
    & (Cosine Similarity)   \\\hline
15   & 0.0002 & 0.1105 & 0.0336 & 0.0117 & 0.0153 & 0.0196 & 0.1012 & 0.8347 \\
30   & 0.0001 & 0.1084 & 0.0321 & 0.0105 & 0.0141 & 0.0184 & 0.0998 & 0.8396 \\
60   & 0.0001 & 0.1072 & 0.0313 & 0.0097 & 0.0130 & 0.0180 & 0.0991 & 0.8421 \\
\bottomrule
\end{tabular}}}}
\end{table}

\subsection{Visualization Analysis}
\begin{figure}[t]
\centering
\subfigure[Real]{\includegraphics[width=.23\textwidth]{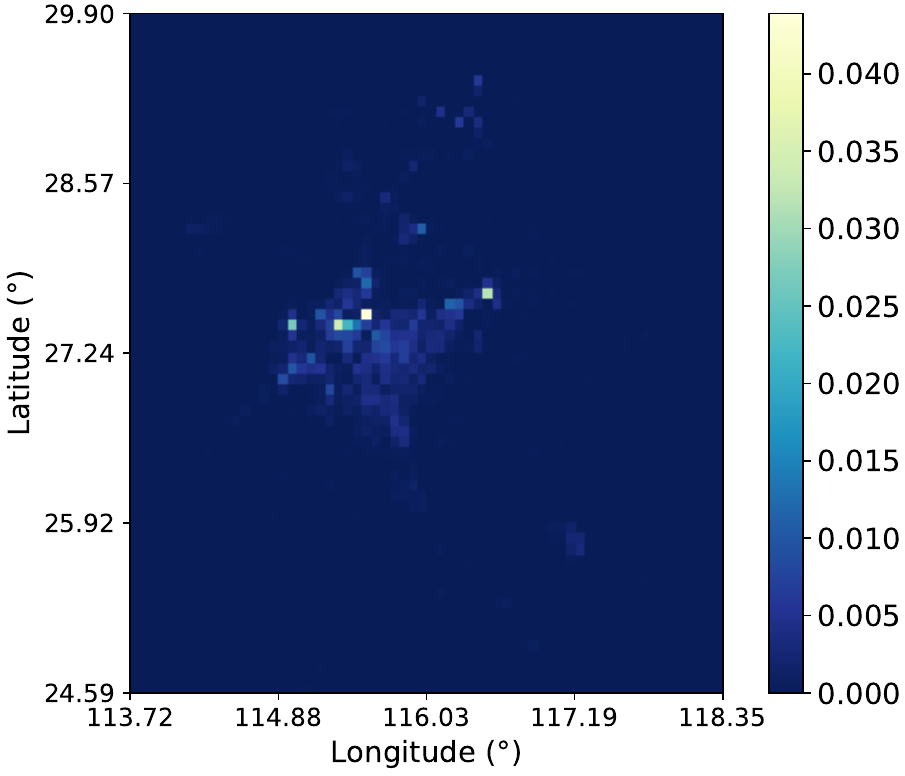}}
\subfigure[MoveSim]{\includegraphics[width=.23\textwidth]{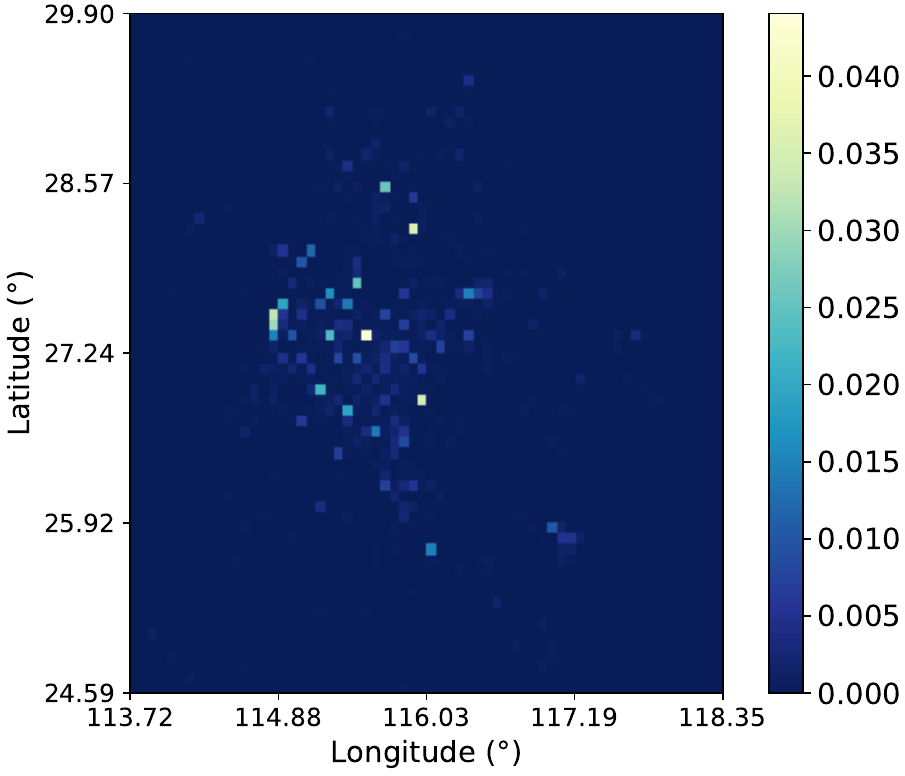}}
\subfigure[PateGail]{\includegraphics[width=.23\textwidth]{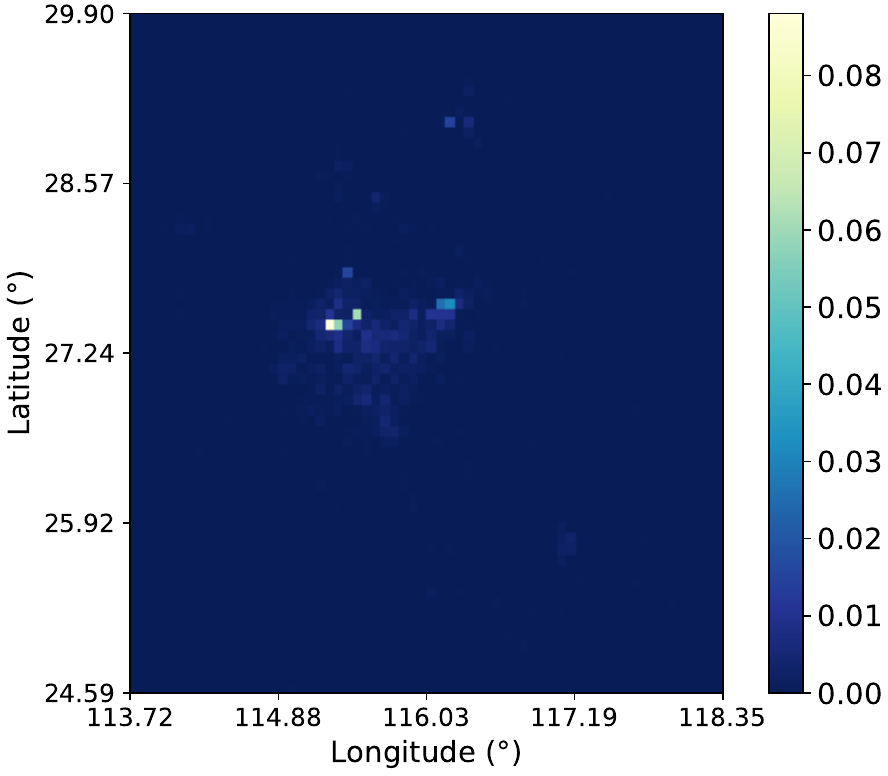}}
\subfigure[Our Model]{\includegraphics[width=.23\textwidth]{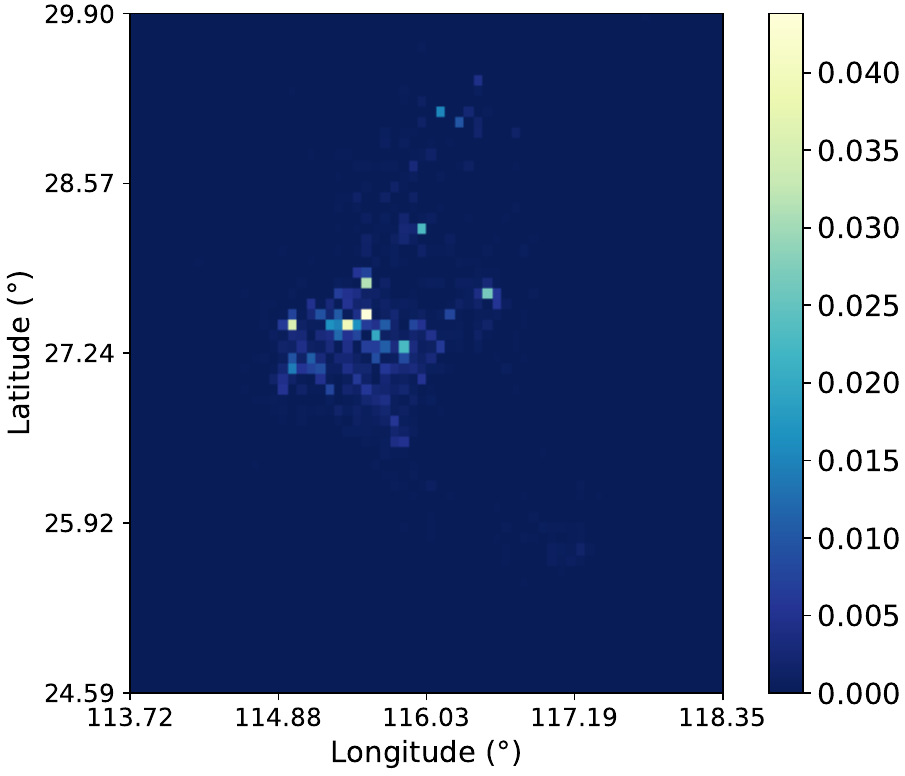}}
\vspace{-10px}
\caption{\revmajor{Comparative analysis of real vs. generated population distributions across a 64$\times$64 grid urban layout.}} 
\label{fig:visual_64}
\end{figure}

\begin{figure}[t]
\centering
\subfigure[Real]{\includegraphics[width=.23\textwidth]{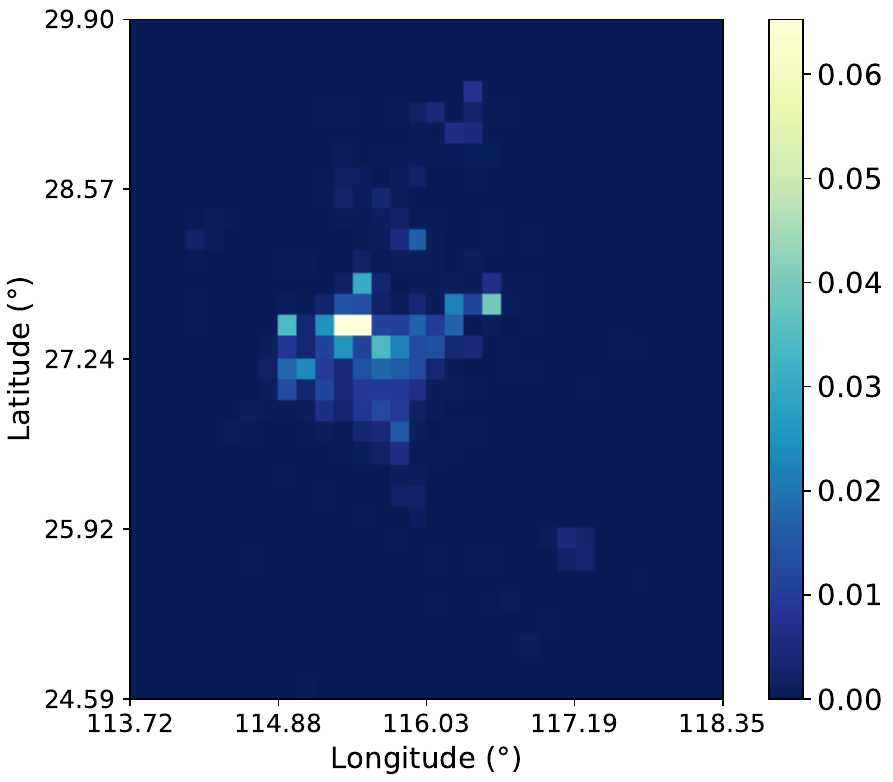}}
\subfigure[MoveSim]{\includegraphics[width=.23\textwidth]{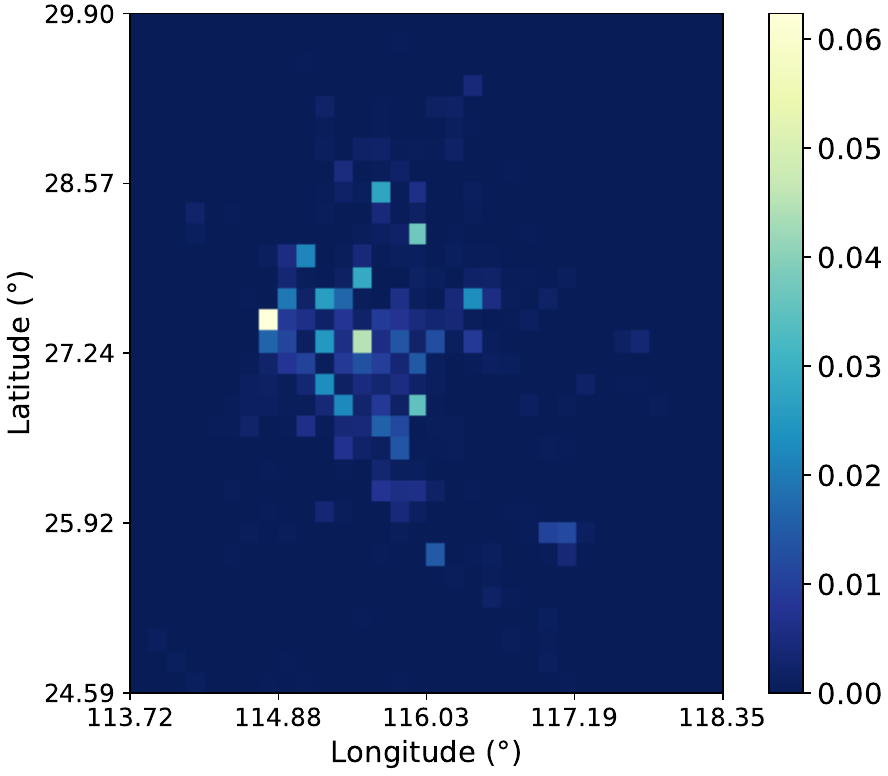}}
\subfigure[PateGail]{\includegraphics[width=.23\textwidth]{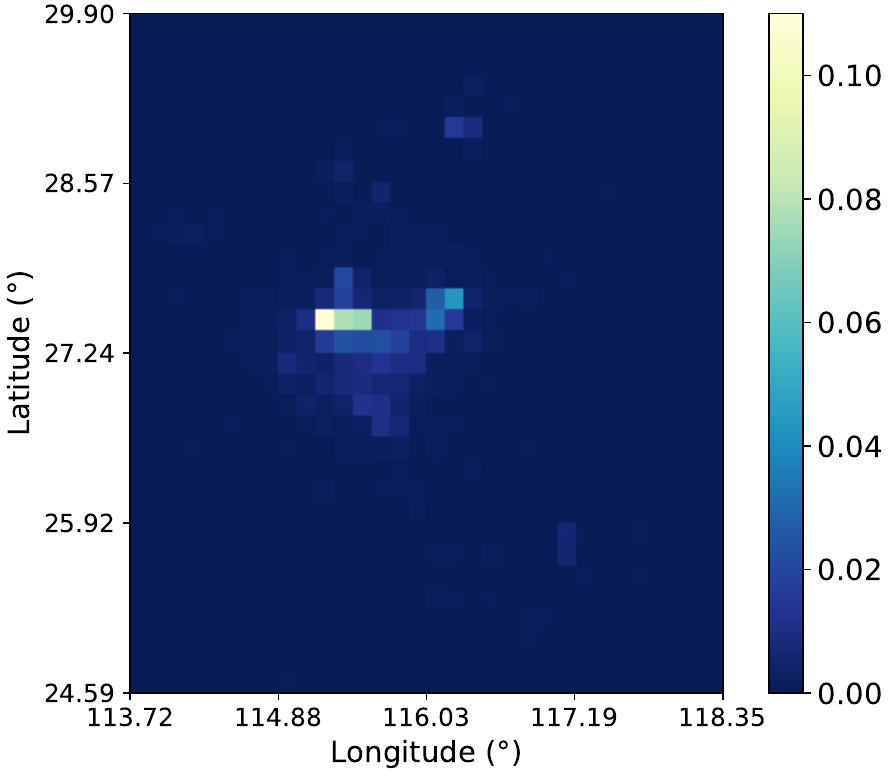}}
\subfigure[Our Model]{\includegraphics[width=.23\textwidth]{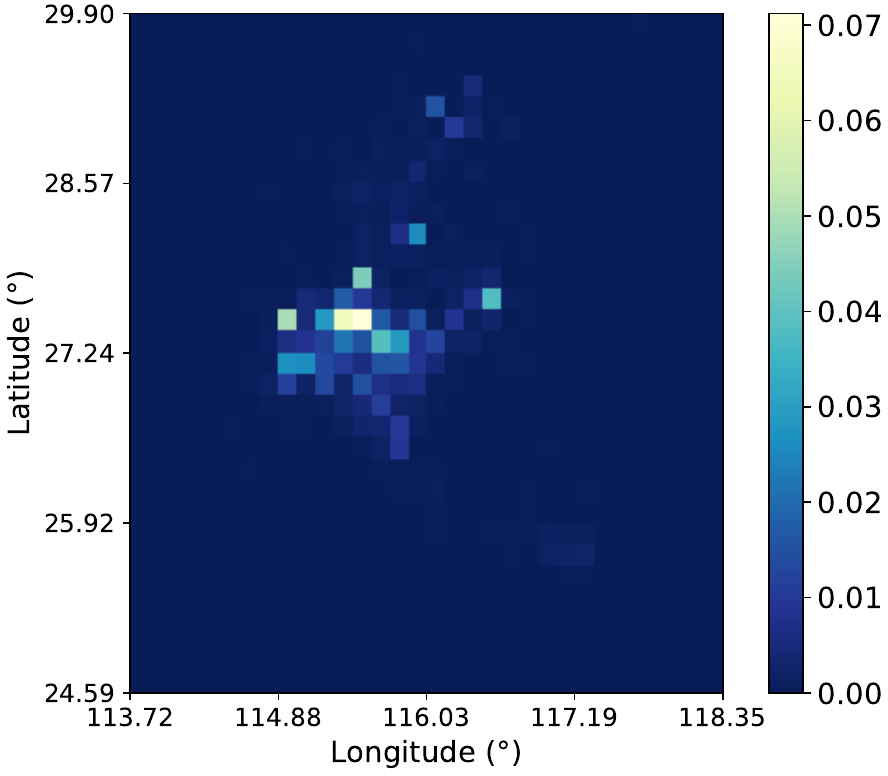}}
\vspace{-10px}
\caption{\revmajor{Comparative analysis of real vs. generated population distributions across a 32$\times$32 grid urban layout.}} 
\label{fig:visual_32}
\end{figure}

\begin{figure}[t]
\centering
\subfigure[Real]{\includegraphics[width=.23\textwidth]{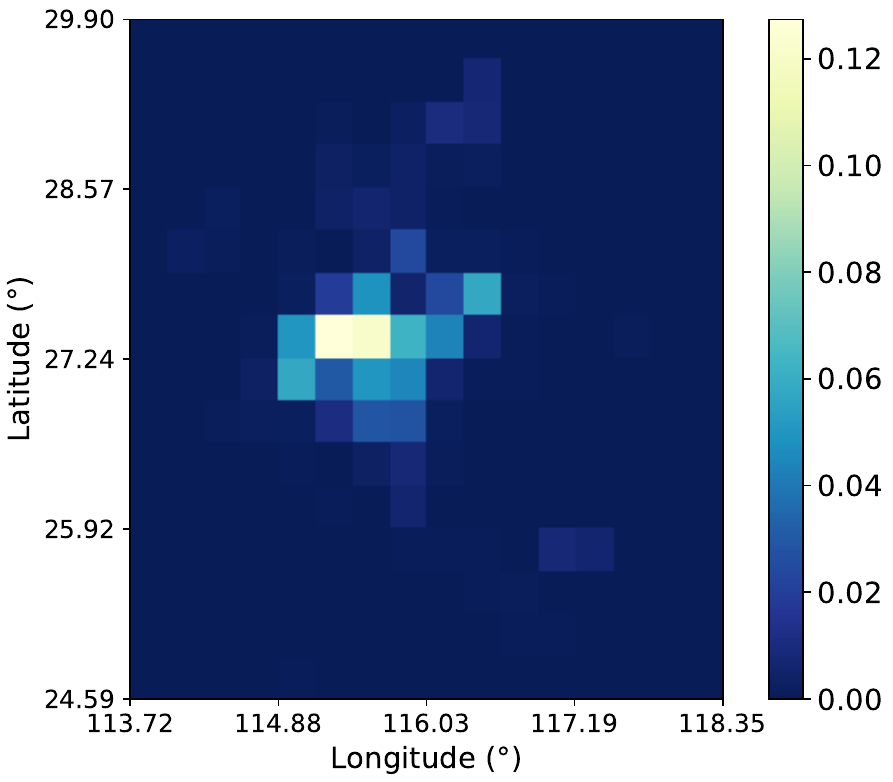}}
\subfigure[MoveSim]{\includegraphics[width=.23\textwidth]{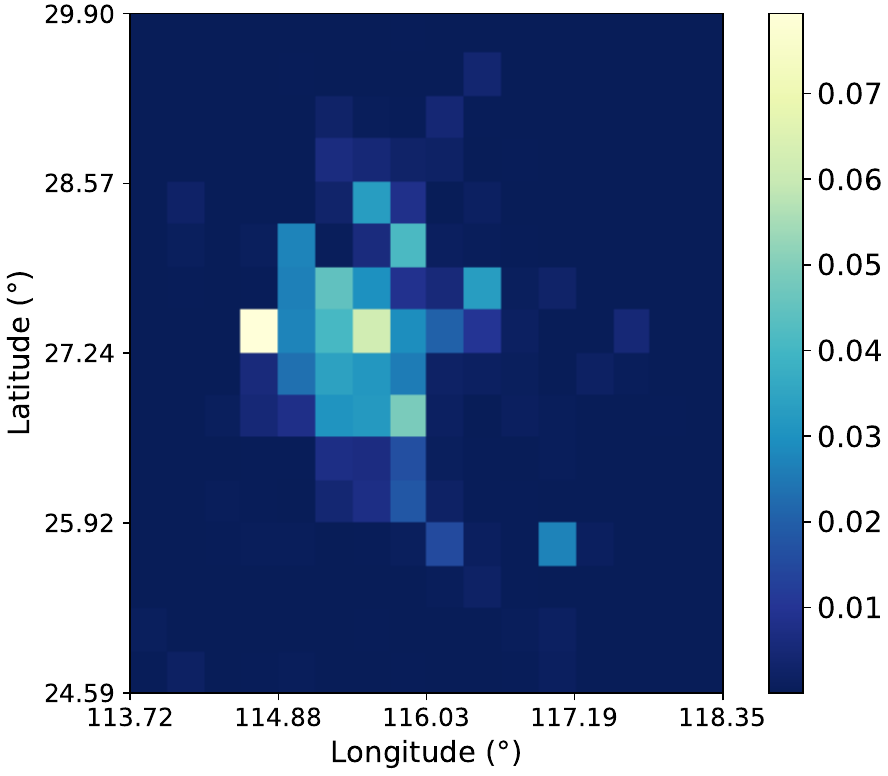}}
\subfigure[PateGail]{\includegraphics[width=.23\textwidth]{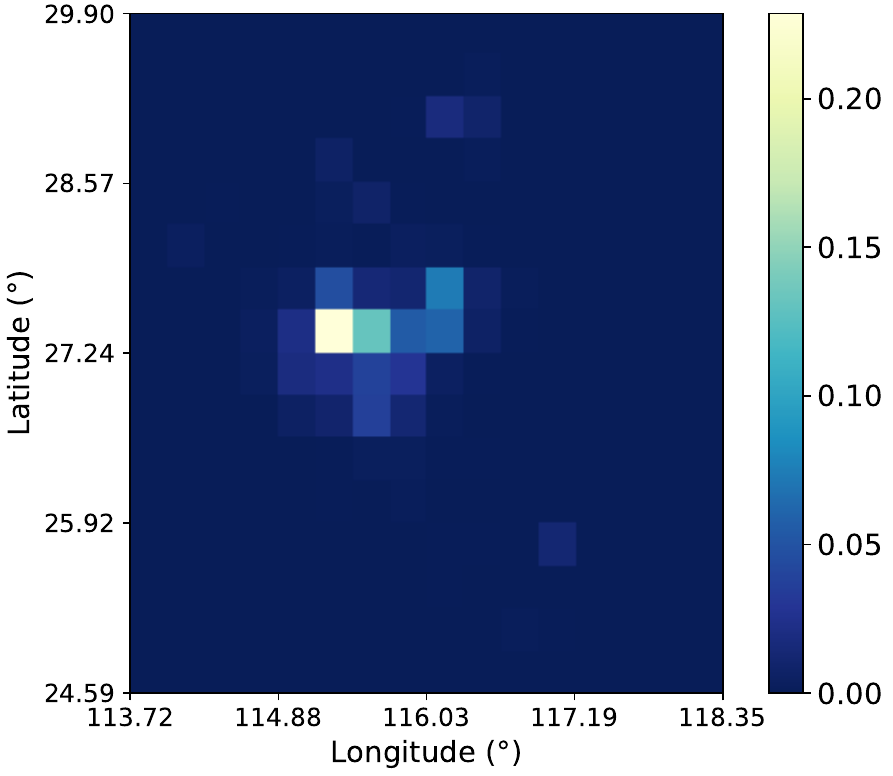}}
\subfigure[Our Model]{\includegraphics[width=.23\textwidth]{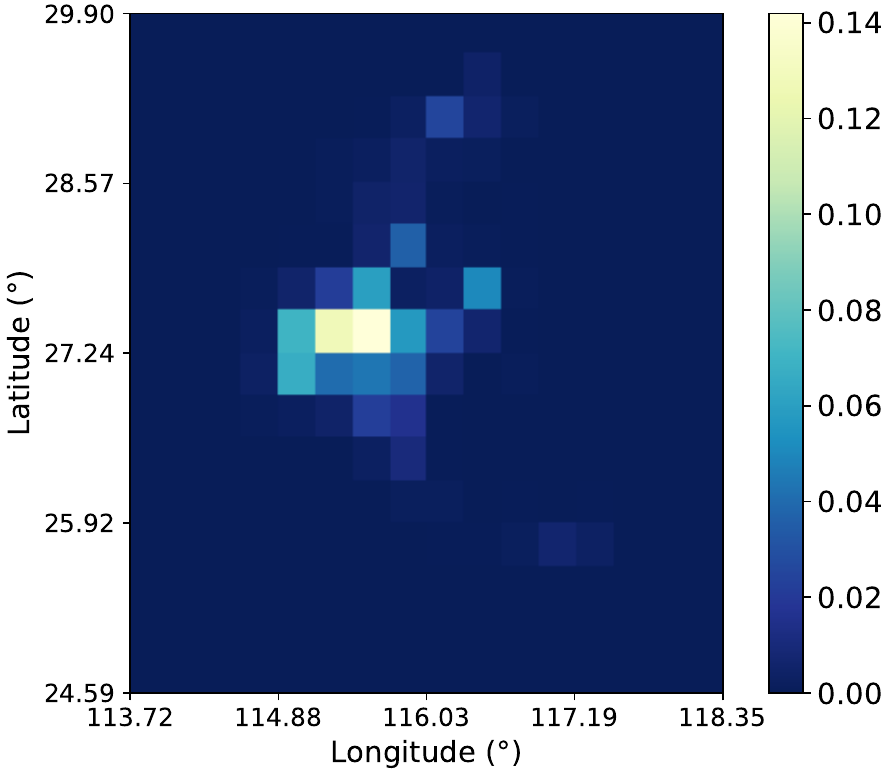}}
\vspace{-10px}
\caption{\revmajor{Comparative analysis of real vs. generated population distributions across a 16$\times$16 grid urban layout.}} 
\label{fig:visual_16}
\end{figure}

\begin{figure}[t]
\centering
\subfigure{\includegraphics[width=0.5\linewidth]{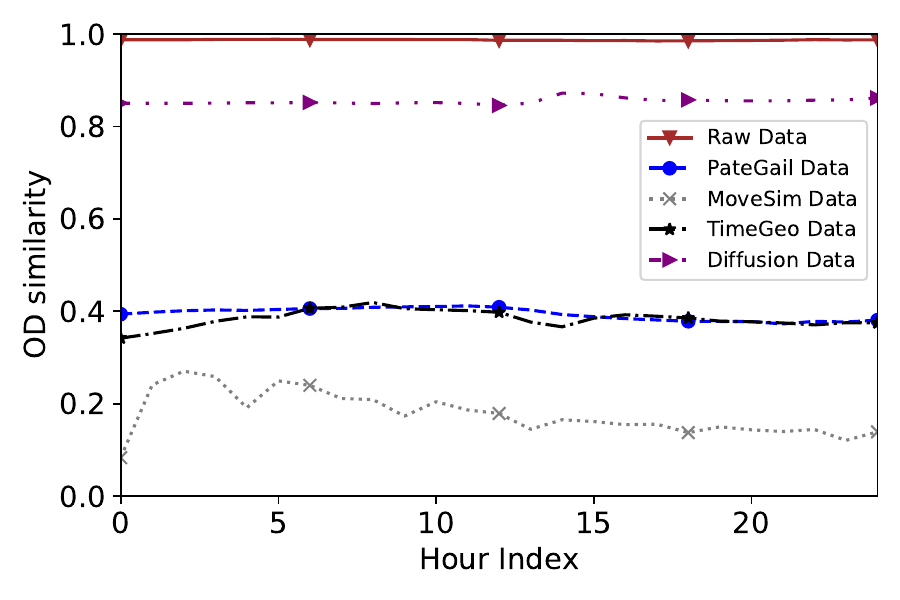}}
\vspace{-10px}
\caption{OD matrix similarity at different hours of the day.} 
\label{fig:od}
\end{figure}

In Figures~\ref{fig:visual_64}, Figures~\ref{fig:visual_32}, and Figures~\ref{fig:visual_16}, we visualize the real and the population distributions generated by various generation methods and present them in heat maps with multiple resolutions. Specifically, we divide the entire city into 16 x 16, 32 x 32, and 64 x 64 grids and then calculate the population distribution in each grid. It can be clearly seen that the trajectories generated by our model are far better at the population distribution level than other generation methods. And it works well at different resolutions, proving that our model maintains a high degree of consistency in the population distribution at both fine and coarse grains. 
Figure~\ref{fig:od} shows average OD matrix similarity in each hour, and we find that our model is closest to the real curve in terms of trend and value.
Combined with the results of several other statistical metrics for mobile trajectories in Table~\ref{tab::performance_geolife} and Table~\ref{tab::performance_mme}, this shows that our model not only maintains the validity and usability of trajectories at the individual level but also highly satisfies the population distribution to generate high-quality trajectories.

%% file: relatedwork.tex
\section{Related work}
\paragraph{Mobility Trajectory Generation.}
Currently, the generation of trajectory data has been an increasing trend in minimizing the difficulty of mobility trajectory data gathering through generating data.  Mobility trajectory generation has been divided into two categories by researchers: knowledge-driven and data-driven~\cite{kong2023mobility}.
Some typical knowledge-driven works are as follows:
The number of jumps within a certain time interval is random in the Continuous-Time Random Walk model (CTRW)~\cite{berkowitz2006modeling}. It can roughly mimic the paths taken by humans. 
In order to accurately simulate human mobility, the Exploration and Preferential Return model~\cite{song2010modelling} makes the assumption that people have a tendency to explore new places and return to previously familiar regions. 
The Social-EPR model~\cite{yabe2023behavioral} assesses changes in mobility behavior during epidemics. It uses a social exploration parameter to measure the probability of visiting locations beyond one's income group when exploring new places.
The main advantages of these methods are their solid theoretical foundation and the transparency of the computational process, but the disadvantage is that they may not be able to handle complex, nonlinear trajectory generation problems.
Some trajectory generation works utilize data-driven methods. 
Liu et al. proposed an initial solution for trajectory generation using Generative Adversarial Networks (GAN)~\cite{liu2018trajgans}. 
Later, some works divide the city map into grids and use the distribution of trajectories between grids to construct trajectories~\cite{zhang2023dp, ouyang2018non}. 
MoveSim~\cite{feng2020learning} further combines prior knowledge and deep learning models by incorporating distance matrices, transition frequency matrices, and functional similarity matrices between locations to enhance the regularity of human mobility.
\rev{TransMob~\cite{he2020human} achieves cross-city trajectory generation by transferring travel patterns from existing cities, while GTG~\cite{wang2025gtg} exploits universal mobility patterns to capture human mobility preferences and generate trajectories.
Long et al. propose the VOLUNTEER framework, which employs a dual VAE architecture to model both user attribute distributions and trajectory behavior distributions, significantly improving the realism and diversity of the generated trajectories~\cite{long2023practical}. 
Zhu et al. introduce a spatiotemporal diffusion probabilistic model for trajectory modeling, effectively combining the diffusion process with the spatiotemporal characteristics of trajectories~\cite{zhu2023difftraj}.}
However, the previous works neglect the guiding significance of dynamic population distribution on trajectories, leading to a lack of practicality in the generated trajectories. Therefore, we propose a trajectory generation method based on diffusion models that incorporates the perception of a dynamic population distribution to fill this gap and enhance the usability of the trajectories.

\paragraph{Diffusion Model.}
First presented by Sohl-Dickstein et al.~\cite{sohl2015deep}, the diffusion model is a probabilistic generative model that was further refined by Ho et al. ~\cite{ho2020denoising} and Song et al. ~\cite{song2020score}. The diffusion model is a new advanced generative model that has been used in a variety of generative tasks, including multimodal learning~\cite{avrahami2022blended, ramesh2022hierarchical}, computer vision~\cite{rombach2022high, li2022srdiff}, and natural language processing~\cite{li2022diffusion, gong2022diffuseq}. 
Recent developments in time series and spatiotemporal modeling have begun to apply potent diffusion models to solve the issue of the widespread use of time series and spatiotemporal data in several crucial real-world applications. TimeGrad~\cite{rasul2021autoregressive} suggested an autoregressive model for multivariate probabilistic time series prediction that makes use of diffusion probabilistic models by estimating the gradients of each time step's data distribution and sampling from it. In order to optimize the diffusion model, CSDI~\cite{tashiro2021csdi} presented a time series imputation technique that takes advantage of the correlations in time data and uses a type of self-supervised training. In order to precisely capture long-term dependencies in time series, SSSD~\cite{alcaraz2022diffusion} combines structured state space models with conditional diffusion models.
\rev{DSTPP~\cite{yuan2023spatio} leverages diffusion point processes to model complex spatiotemporal joint distributions, enabling general spatiotemporal prediction. TrajGDM~\cite{chu2024simulating} is a trajectory generation framework using the diffusion model to encapsulate the universal mobility pattern inside a trajectory dataset. Hu et al. propose a diffusion model that integrates spatiotemporal features and urban contexts for predicting travel destinations~\cite{liu2025spatio}. Liu et al. combine a Transformer with diffusion models for spatiotemporal probabilistic forecasting of wind speed~\cite{hu2024predicting}. Wei et al. introduce Diff-RNTraj, a structure-aware diffusion model that generates trajectories constrained by real road networks\cite{wei2024diff}. Liu et al. propose an embedding-space conditional diffusion model that denoises in a latent space to recover missing human trajectory points~\cite{liugenerative}.}
However, previous work has explored diffusion models to a limited extent in generating spatiotemporal human mobility data, especially for human mobility generation considering population distributions. Our framework uses a diffusion model to generate trajectories that conform to dynamic population distributions, improving the realism and usefulness of the generated trajectories.